\pdfoutput=1
\documentclass[11pt]{article}
\usepackage[dvipsnames]{xcolor}         % colors

\usepackage[]{ACL2023}

\usepackage{times}
\usepackage{latexsym}

\usepackage[T1]{fontenc}
\usepackage[utf8]{inputenc}

\usepackage{microtype}

\usepackage{inconsolata}
\usepackage[T1]{fontenc}    % use 8-bit T1 fonts
\usepackage{hyperref}       % hyperlinks
\usepackage{url}            % simple URL typesetting
\usepackage{booktabs}       % professional-quality tables
\usepackage{amsfonts}       % blackboard math symbols
\usepackage{nicefrac}       % compact symbols for 1/2, etc.

\usepackage[shortlabels]{enumitem}
\usepackage{subcaption}
\usepackage{wrapfig}
\usepackage{graphicx}
\usepackage{tikz}
\usetikzlibrary{positioning}
\usepackage{xspace}
\usepackage{tcolorbox}
\usepackage{cleveref}
\usepackage{multirow}
\usepackage{makecell}

\usepackage{color-edits}

\addauthor{gn}{magenta}
\newcommand\bitsmall{\fontsize{9.5}{11}\selectfont}

\title{In-Context Learning with Long-Context Models: \\ An In-Depth Exploration}

\author{Amanda Bertsch\textsuperscript{$\gamma$}\\
\bitsmall{\texttt{abertsch@cs.cmu.edu}}
\And
Maor Ivgi \textsuperscript{$\tau$} \\
\bitsmall{\texttt{maor.ivgi@cs.tau.ac.il}}
\And
Emily Xiao\textsuperscript{$\gamma$}\\
\bitsmall{\texttt{emilyx@cs.cmu.edu}}
\AND
Uri Alon\textsuperscript{$\gamma$}\thanks{\hspace{0.5em} Now at Google DeepMind} \\
\bitsmall{\texttt{urialon@cs.cmu.edu}}
\And
Jonathan Berant\textsuperscript{$\tau$} \\
\bitsmall{\texttt{joberant@cs.tau.ac.il}}
\And
Matthew R. Gormley\textsuperscript{$\gamma$} \\
\bitsmall{\texttt{mgormley@cs.cmu.edu}}
\And
Graham Neubig\textsuperscript{$\gamma$} \\
\bitsmall{\texttt{gneubig@cs.cmu.edu}} 
\AND 
\textsuperscript{$\gamma$} Carnegie Mellon University \quad
\textsuperscript{$\tau$} Tel Aviv University \\
}

\newcommand{\banking}{Banking-77\xspace}
\newcommand{\clinic}{Clinic-150\xspace}
\newcommand{\trec}{TREC\xspace}
\newcommand{\trecfine}{TREC-fine\xspace}
\newcommand{\nlu}{NLU\xspace}
\newcommand{\samsum}{SAMSum\xspace}
\newcommand{\longllama}{Llama2-80k\xspace}
\newcommand{\togetherllama}{Llama2-32k\xspace}

\begin{document}

\maketitle

\begin{abstract}
As model context lengths continue to increase, the number of demonstrations that can be provided in-context approaches the size of entire training datasets. We study the behavior of in-context learning (ICL) at this extreme scale on multiple datasets and models. We show that, for many datasets with large label spaces, performance continues to increase with thousands of demonstrations. 
We contrast this with example retrieval and finetuning: example retrieval shows excellent performance at low context lengths but has diminished gains with more demonstrations; finetuning is more data hungry than ICL but can exceed long-context ICL performance with additional data. 
We use the ICL setting to study several properties of both in-context learning and long-context models.
We show that long-context ICL is less sensitive to random input shuffling than short-context ICL, that grouping of same-label examples negatively impacts performance, and that the performance boosts do not arise from cumulative gain from encoding many examples together.
We conclude that long-context ICL can be an effective tool, and may not require long-context for encoding the demonstration set at all.
\footnote{Data and code are available at \url{https://github.com/abertsch72/long-context-icl}}
\end{abstract}

\section{Introduction}

\begin{figure}
  \centering
  \centering
  \includegraphics[width=0.9\linewidth]{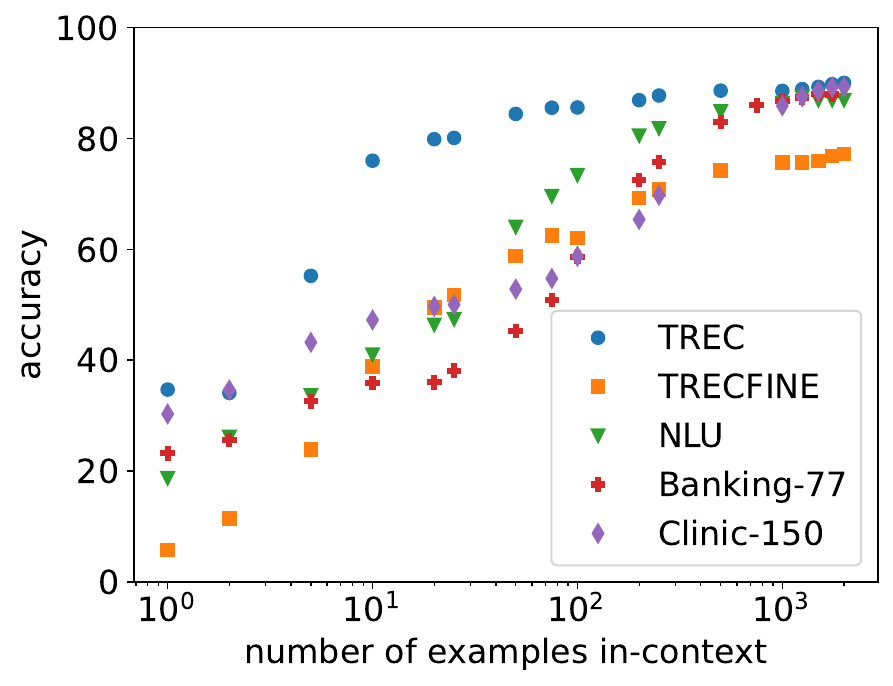}
  \label{fig:random-all}
\caption{The performance increases with more demonstrations far beyond the context window of the base Llama-2. Results are on \citet{fu2024data}'s long-context finetuned Llama-2-7b model, using a context of up to 80K tokens. 
}
\label{fig:perf-summary}
\end{figure}

When a few examples are provided in-context, large language models can perform many tasks with reasonable accuracy. While questions remain about the exact mechanism behind this phenomena \citep{min-etal-2022-rethinking, vonoswald2023transformers}, this paradigm of \emph{in-context learning} (ICL) has seen widespread adoption in both academic and industry applications, thanks to its ease of implementation, relatively small computational cost, and ability to reuse a single model across tasks.

However, most work has focused on models where the maximum number of demonstrations is severely limited by context length. As more and more methods are developed to adapt language models to extreme context lengths (\citet{gemini1.5, fu2024data}, \textit{inter alia}), in-context learning over large quantities of data becomes a potential alternative to finetuning. The properties of ICL in this regime are not well-understood; and as the cost of inference over many thousands of  tokens can be steep, the efficiency and performance tradeoff between many-shot ICL and finetuning on the same data is complex. 

\begin{table*}[h]
\begin{center}
\begin{tabular}{llcccc}
\toprule
\textbf{Dataset} &  \textbf{Domain} & \textbf{\# Labels} & \multicolumn{1}{c}{\textbf{\makecell[c]{Avg \\demo \\ length}}} & \textbf{\makecell[c]{Training \\ set size}} &  \textbf{Example outputs}  \\
\midrule
\trec & questions & 6 & 22.7 & 5,452 & \makecell[c]{location  / entity \\ } \\[2pt]
\trecfine & questions & 50 & 23.7  & 5,452 & \makecell[c]{abbreviation expansion /  location city}  \\[2pt]
\nlu & conversational & 68 & 20.7  & 19,286 &  \makecell[c]{takeaway query / iot hue light up} \\[2pt]
\banking & financial & 77 & 27.4   & 10,003 &    \makecell[c]{top up failed / lost or stolen card}\\[2pt]
\clinic & multiple & 151 & 22.3 & 15,250 & \makecell[c]{rollover 401k / meal suggestion \\} \\[2pt]
\samsum & conversational & n/a & 167.9 & 14,732 & \makecell[c]{John will buy the goat cheese Tracy \\ liked, milk, a couple of grainy rolls,\\ and tissues.} \\[2pt]

\bottomrule
\end{tabular}
\end{center}
\caption{The datasets we consider in this work span diverse label spaces and domains. The average demonstration length is the average combined length of input, output, and formatting tokens per demonstration provided in the context window; outliers in the top 1\% for demonstration length are discarded and not reflected in these statistics.}
\label{tab:dataset-stats}
\end{table*}

We conduct a systemic study of long-context in-context learning. Namely, we consider: a) the performance of prompting the base model naively, b) retrieving examples to use in-context for each test example, c) finetuning the base model (both full and parameter-efficient finetuning), and d) using models trained to adapt to longer contexts. 
 Performance continues to increase past 2000 demonstrations (see Figure~\ref{fig:perf-summary}), approaching and sometimes \emph{exceeding} the performance of models finetuned on thousands of examples from the same dataset (\S~\ref{sec:results}).

 We find that, as the number of demonstrations in-context increases to extreme values, the behavior of ICL shifts (\S~\ref{sec:analysis}).
In-context learning becomes less sensitive to example order, and the benefits of retrieval over using a random set of demonstrations diminishes --- allowing the use of a single set of demonstrations, encoded once through the model and cached, rather than re-encoding a custom set of demonstrations for each example. We demonstrate that long-context ICL is strongly impacted by grouping examples of the same label. We also find that the effectiveness of long-context ICL is not dependent on long-range attention in the demonstration set-- encoding demonstrations with local attention and using global attention only for the test example recovers nearly the same performance (\S~\ref{sec:why}). Our work furthers the understanding of in-context learning and shows that %, in some data regimes,
long-context ICL is a strong alternative to retrieval and finetuning.

\section{Experimental setup}

We consider 5 classification datasets: \trec \citep{hovy-etal-2001-toward}, \trecfine \citep{hovy-etal-2001-toward}, \nlu \citep{nlu}, \banking \citep{casanueva-etal-2020-efficient}, and \clinic \citep{clinic150}; and 1 generation dataset: \samsum \citep{Gliwa_2019}.
Table~\ref{tab:dataset-stats} contains summary statistics for each dataset, and Appendix~\ref{appendix:prompts} shows additional description for each dataset.

We compare ICL performance across several long- and short-context models, including variants of Llama-2 with 4k  \citep{touvron2023llama}, 32k  \citep{TogetherAI_2023}, and 80k \citep{fu2024data} context windows, Mistral-7b-v0.2 \citep{jiang2023mistral}, and Qwen 2.5-7B \citep{qwen2.5}. For more details on models, see Appendix~\ref{appendix:extras}.

\begin{table*}[t]
\begin{center}
\begin{tabular}{llllll}
\toprule
\multicolumn{1}{c}{\bf Dataset}  &\multicolumn{1}{c}{\bf Llama2}  &\multicolumn{1}{c}{\bf Llama2-32k}  &\multicolumn{1}{c}{\bf Llama2-80k}  & {\bf Mistral}  & {\bf Qwen2.5}\\ 
\midrule 
Randomly selected \\
\midrule
\trec & \textbf{82.32} / 80.52 & \textbf{93.12} / \textbf{93.12} & \textbf{90.04} / \textbf{90.04} & \textbf{87.28} / 85.00  & \textbf{94.68} / 94.40 \\
\trecfine & \textbf{61.40} / \textbf{61.40} & \textbf{75.56} / 75.08 & \textbf{77.20} / \textbf{77.20} & \textbf{72.68} / 70.48 & \textbf{83.40} / 81.24  \\
\nlu & \textbf{76.88} / \textbf{76.88} & \textbf{85.04} / 85.00 & \textbf{87.52} / 86.92 & \textbf{86.44} / \textbf{86.44} & \textbf{88.64} / \textbf{88.64} \\
\banking & \textbf{56.36} / \textbf{56.36} & \textbf{82.44} / \textbf{82.44} & \textbf{88.08} / 87.96 & \textbf{86.76} / 86.68  & \textbf{88.60} / 87.96 \\
\clinic & \textbf{60.92} / \textbf{60.92} & \textbf{84.40} / \textbf{84.40} & \textbf{89.32} / \textbf{89.32} & \textbf{90.56} / \textbf{90.56} &  \textbf{93.16} / 92.76  \\
\samsum & \textbf{79.83} / \textbf{79.83} &  \textbf{81.65} / 81.42 & \textbf{81.17} / \textbf{81.17}  & \textbf{81.78} / \textbf{81.78} & \textbf{82.01} / 81.86  \\
\toprule
\midrule
BM25 Retrieval \\
\midrule
\trec & \textbf{90.80} / 85.64 & \textbf{94.84} / 94.64 & \textbf{94.28} / 92.68 & \textbf{90.80} / \textbf{90.80} & \textbf{95.60} / 95.20  \\
\trecfine & \textbf{78.80} / \textbf{78.80} & \textbf{83.88} / 81.12 & \textbf{83.92} / 81.36 & \textbf{80.80} / 79.60 & \textbf{88.00} / \textbf{88.00} \\
\nlu & \textbf{90.00} / 88.40 & \textbf{89.80} / \textbf{89.80} & \textbf{89.64} / 89.52 & \textbf{90.40} / 89.20 & \textbf{90.40} / \textbf{90.40}  \\
\banking & \textbf{93.20} / 92.40 & \textbf{94.32} / \textbf{94.32} & \textbf{94.00} / 92.96 & \textbf{93.20} / \textbf{93.20} & \textbf{92.80} / 91.60  \\
\clinic & \textbf{87.60} / \textbf{87.60} & \textbf{89.84} / \textbf{89.84} & \textbf{93.76} / \textbf{93.76} & \textbf{93.20} / 92.40 & \textbf{95.20} / 93.20 \\
\samsum & \textbf{79.98} / \textbf{79.98} &  \textbf{81.40} / 81.19 & \textbf{80.68} / \textbf{80.68}  & \textbf{81.33} / \textbf{81.33} & \textbf{81.90} / \textbf{81.90}  \\
\bottomrule
\end{tabular}
\end{center}
\caption{For all datasets, performance of ICL continues to increase with additional demonstrations. These results are the best accuracy (left) and accuracy at maximum data (right) for each model on the classification tasks, and the same with BERTScore for \samsum.  Bold indicates the best performance for that model/dataset pair. }
\label{tab:results-main}

\end{table*}
\paragraph{Constrained decoding}
For each classification dataset, we use \emph{constrained decoding} to only produce valid labels as output.
Note that, without constrained decoding, these models may produce invalid labels in the few-shot regimes (see Appendix~\ref{appendix:unconstrained}). 
For finetuning, we use a classification head; thus no invalid outputs can be produced.

\paragraph{Evaluation}
Following prior work \citep{zhao2021calibrate, lu-etal-2022-fantastically, han2022prototypical, Ratner2022ParallelCW}, we subsample 250 examples from the test set of each dataset. 
 We release the subsampled test set and full prediction outputs for each experiment in the project repository.
We evaluate on each classification dataset with accuracy and macro-F1
; as the trends for the metrics are very similar, we report accuracy (the more common metric) in the paper. We evaluate on \samsum with BERTScore \cite{zhang2020bertscoreevaluatingtextgeneration} and confirm that we see similar trends in ROUGE \citep{lin-2004-rouge}, as measured using the \texttt{rouge\_scorer} package.\footnote{\url{https://pypi.org/project/rouge-score/}}

\section{Long-context ICL}
\label{sec:results}
We consider four common methods for using a large dataset. 
    
\subsection{Compared settings}

\begin{figure*}
  \centering
\begin{subfigure}{.5\textwidth}
  \centering
  \includegraphics[width=0.9\linewidth]{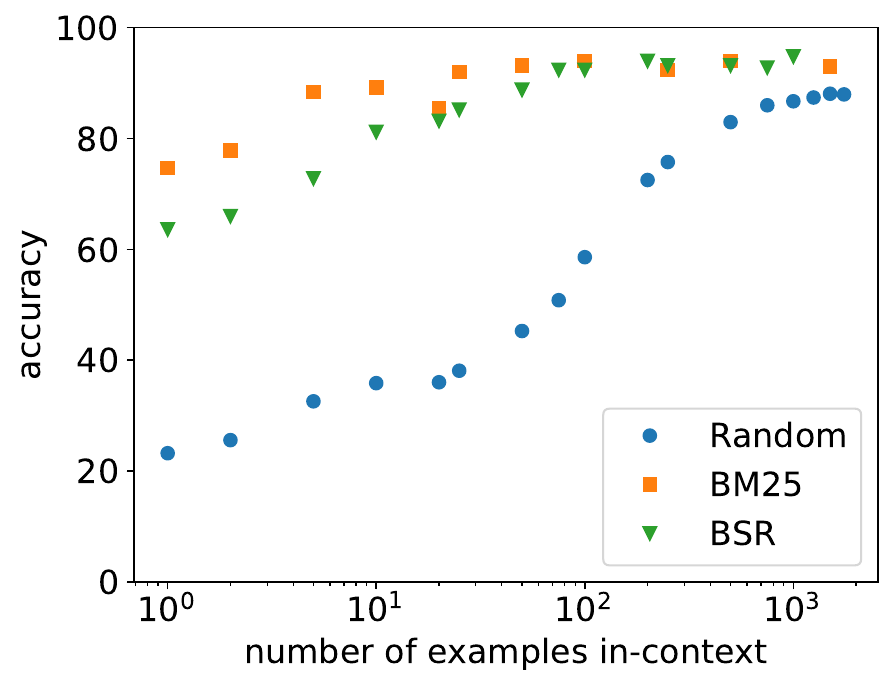}
  \caption{\banking}
  \label{fig:banking-ret-comp}
\end{subfigure}%
\begin{subfigure}{.5\textwidth}
  \centering
  \includegraphics[width=0.9\linewidth]{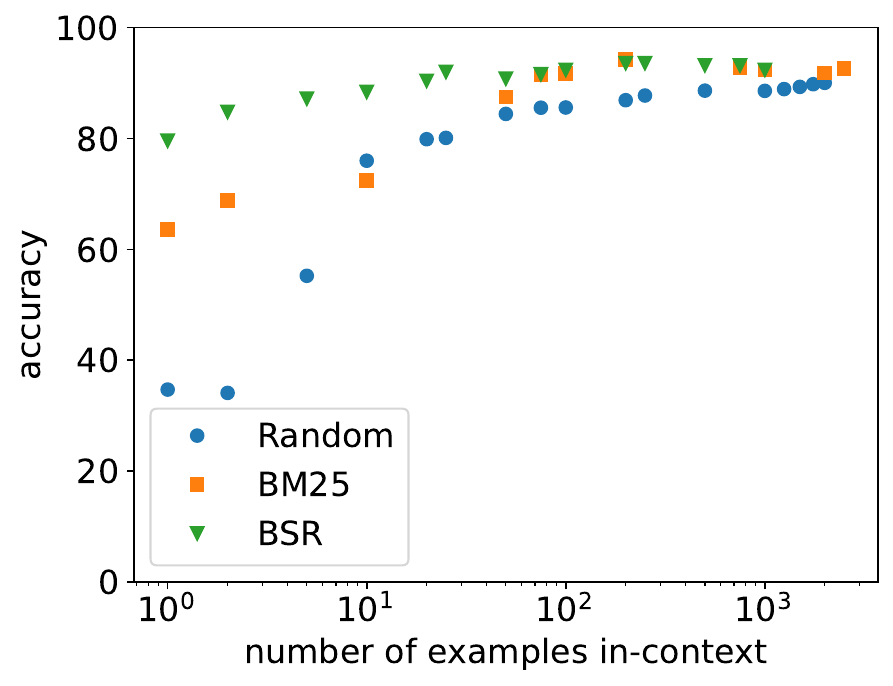}
  \caption{\trec}
  \label{fig:trec-ret-comp}
\end{subfigure} \\
\caption{Comparing three selection methods-- random selection, BM25, and BERTScore-Recall (BSR) on two representative datasets. At smaller numbers of demonstrations in-context, BM25 and BSR have differing performance, and the best retriever is dataset-specific; at larger demonstration counts, the two become indistinguishable. Both generally outperform random selection.}.
\label{fig:compare-methods}
\end{figure*}

\paragraph{Random sampling ICL} We use 10 random shuffles of the training dataset, averaging the results across these shuffles. Across models and across varying numbers of demonstrations in-context, we draw the first $n$ examples from each shuffle. In this setting, the encoding of demonstrations can be performed once and cached.

\paragraph{Retrieval ICL} A strong alternative for in-context learning is to retrieve a relevant subset of examples as demonstrations for each test set example. Prior work has found that, in some scenarios, retrieval of good examples can make the difference from near-zero to high test accuracy \citep{levy-etal-2023-diverse}.
We considered two possible retrievers for this setting: BM25 \citep{bm25} and BERTScore-Recall \citep{gupta2023coveragebasedexampleselectionincontext}. For both, we retrieve the most relevant demonstrations by comparing the test input text to the full demonstration texts. For BM25, we remove stopwords; when doing $k$-shot prompting, if less than $k$ examples are retrieved by the retriever,\footnote{This occurs when there are less than $k$ examples with any (non-stopword) overlap with the test example.} we randomly sample additional examples until we reach $k$.

\paragraph{Finetuning} 
We finetune Llama2-7b with a classification head on varying amounts of data from each dataset with several random seeds, and plot performance at convergence on the same held-out test data. 
We initialize the classification head from the parameters of the pretrained language modeling head by subsampling the values of the first token of each label; this creates a better-than-random initialization for finetuning.   
We perform both full finetuning and LoRA \cite{hu2021lora} finetuning using this setup; for more details on the finetuning procedures, see Appendix~\ref{appendix:finetuning}. 

\subsection{In-context results}

\paragraph{Scaling up ICL to many examples leads to surprisingly strong results}
Figure~\ref{fig:perf-summary} and Table~\ref{tab:results-main} show the performance of models in the ICL settings. Scaling up from 10 to 1000 demonstrations results in accuracy gains of up to \emph{50.8} points (and an average of 36.8 points across 5 datasets for \longllama).

\paragraph{Longer context lessens the importance of carefully selecting in-context examples} 

Retrieving relevant examples for each test set example far outperforms using a randomly selected subset in the short-context regime.
This is true even if the order of retrieved examples is shuffled (rather than ordered by relevance).\footnote{We perform three random shuffles of the BM25 retrieved inputs and test for difference in distribution from the original results; this does not significantly change performance for any dataset (2-sided t-test, $p < 0.05$).}
However, adding additional examples does continue to slightly improve performance; this is especially surprising for the BM25 retriever because, after all examples with non-trivial lexical overlap are retrieved, remaining examples are randomly selected. 

While retrieval continues to outperform random selection, the importance of the selection strategy diminishes with additional examples. On some datasets (e.g. \trec, in Figure~\ref{fig:trec-ret-comp}), BERTScore-Recall outperforms BM25 for short context ICL; on other datasets (e.g. \banking, in Figure~\ref{fig:banking-ret-comp}), the inverse is true. But on all datasets, the performance difference between the two retrieval methods diminishes with larger $k$, so that BM25 and BSR have nearly identical performance at long-context ICL. Because of this, we report only BM25 in the remainder of the analysis.

As the performance difference between individual retrievers diminishes, so does the performance difference between retrieval and random selection of demonstrations. On \banking, the dataset where retrieval is most beneficial, the performance gain from BM25 retrieval drops from 51.5 points at 1-shot ICL to 4.9 points at 1500-shot ICL. 
This is compelling because it is more computationally efficient (but less effective) to encode a single random set of demonstrations and cache them, rather than retrieving and re-encoding a custom set of demonstrations for each inference example. In the longest context regime we consider, using a single randomly selected set of examples is feasible; the performance penalty for doing so is never more than 5 points, and as low as 1.8 points (in 2000-shot ICL on \trec).

\paragraph{Long-context ICL is also effective for generation.} While we primarily focus on classification tasks because of the relative ease of evaluation, we do consider \samsum, a text summarization task, for our ICL experiments in Table~\ref{tab:results-main}. While the number of demonstrations possible in the same context length is much smaller, due to the increased lengths of both inputs and outputs, we observe increased performance with additional demonstrations up to at least 250-shot ICL. Retrieval seems less helpful in this setting, with retrieval sometimes underperforming random selection.

\begin{figure*}[h]
  \centering
\begin{subfigure}{.5\textwidth}
  \centering
  \includegraphics[width=0.9\linewidth]{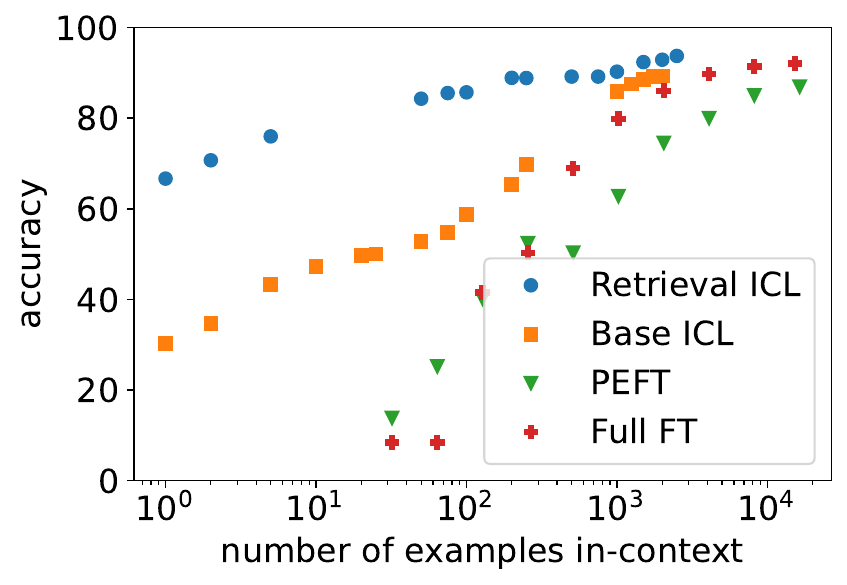}
  \caption{Clinic-150}
  \label{fig:clinic-comp}
\end{subfigure}%
\begin{subfigure}{.5\textwidth}
  \centering
  \includegraphics[width=0.9\linewidth]{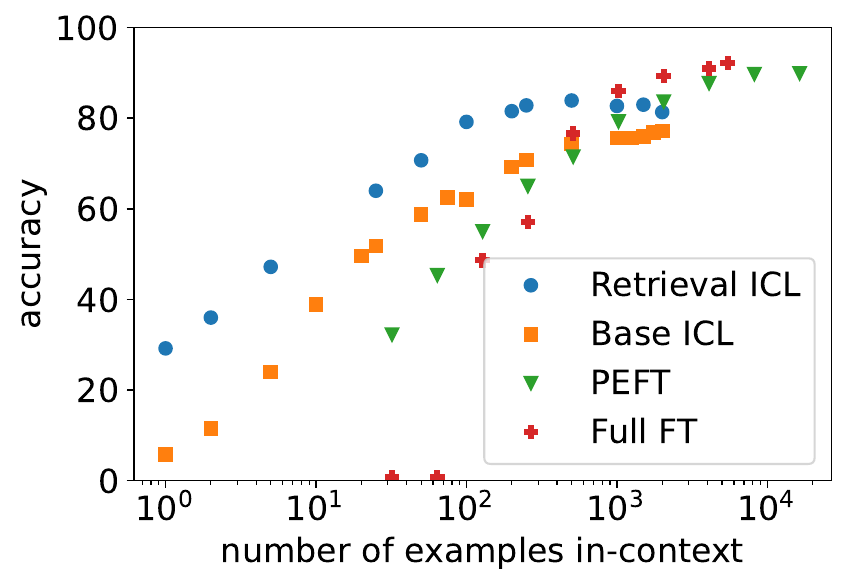}
  \caption{Trecfine}
  \label{fig:trecfine-comp}
\end{subfigure} \\
\caption{Comparing BM25 retrieval ICL, random selection ICL, and two types of finetuning on two representative datasets. Finetuning sometimes, but not always, exceeds ICL at high numbers of demonstrations. Note that, while retrieval ICL uses the listed number of examples in context, it assumes access to the larger test set to draw examples from \citep{perez2021true}. See Appendix \ref{appendix:all-figs} for results on other datasets.}.
\label{fig:compare-methods}
\end{figure*}

\subsection{Comparison with finetuning} 

While we have demonstrated that in-context learning with hundreds or thousands of examples is effective, this amount of data is also appropriate for finetuning a model. 
Finetuning has higher upfront cost but allows for reduced inference-time cost. 
We compare in-context learning with full finetuning and the popular parameter-efficient finetuning (PEFT) strategy LoRA \citep{hu2021lora}.

\paragraph{Finetuning is (slightly) more data-hungry than ICL} When a relatively small set of examples is available, ICL generally outperforms LoRA finetuning on the same model.\footnote{Note that some prior works have showed strong PEFT performance in the few-example setting on different tasks; see \Cref{sec:related-work} for more discussion.} 
For most datasets, LoRA finetuning performance never exceeds long-context ICL performance even with additional examples (e.g. \Cref{fig:clinic-comp}); however, for most datasets, full finetuning with drastically more examples than fit in-context yields the highest performance.
Generally, the datasets with larger label spaces show the least strong finetuning performance, likely because these are more open-ended classification problems and require more data to train the classifier; on the dataset with the most labels, \clinic, neither LoRA finetuning nor full finetuning ever outperforms ICL at the same number of examples.

In a setting with unlimited training data available, then, finetuning is clearly advantageous over ICL for any model with a fixed maximum context length. 
Finetuning also offers dramatically reduced inference costs for similar performance; thus, finetuning on 4096 examples may still be preferable to prompting with 1000 if efficiency of inference is a major priority.\footnote{With the caveat that serving several task-specific models may be more expensive than serving one general-purpose model with customized ICL prompts; the actual best cost/efficiency tradeoff in any downstream setting is of course dependent on the needs of the deploying organization.} This is because, even if demonstration encodings can be cached across inference examples, cross-attention to long context is costly. 

\section{Properties of long-context ICL}
\label{sec:analysis}

\begin{figure*}[htbp]
    \centering
    \begin{minipage}[t]{0.47\textwidth}
        \centering
        \includegraphics[width=0.9\linewidth]{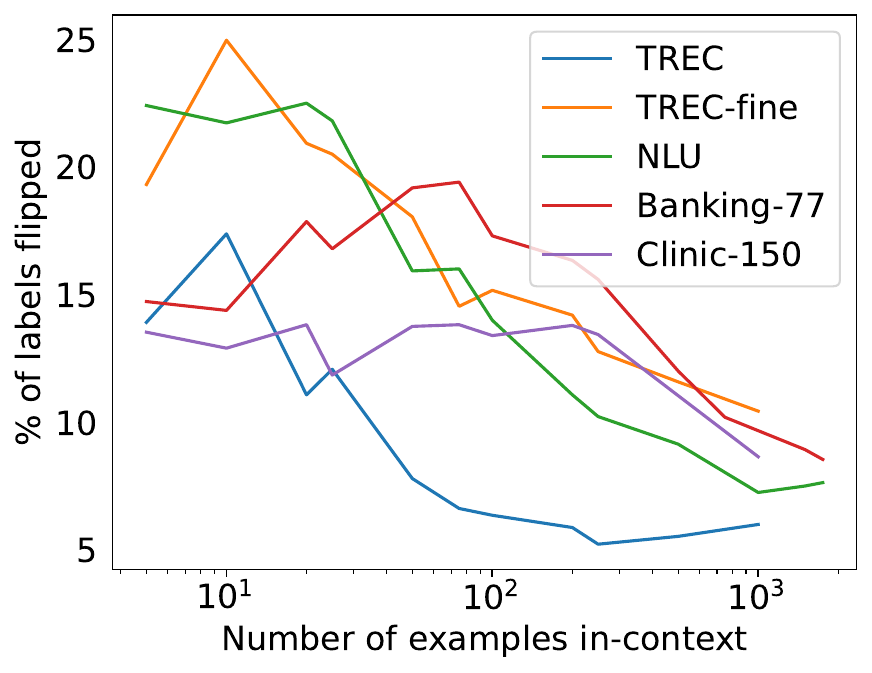} % Adjust width as needed
  \caption{The impact of (randomly) reordering examples in-context decreases with additional demonstrations.}
  \label{fig:reording-demonstrations}
    \end{minipage}
    \hfill
    \begin{minipage}[t]{0.47\textwidth}
        \centering
        \includegraphics[width=0.9\linewidth]{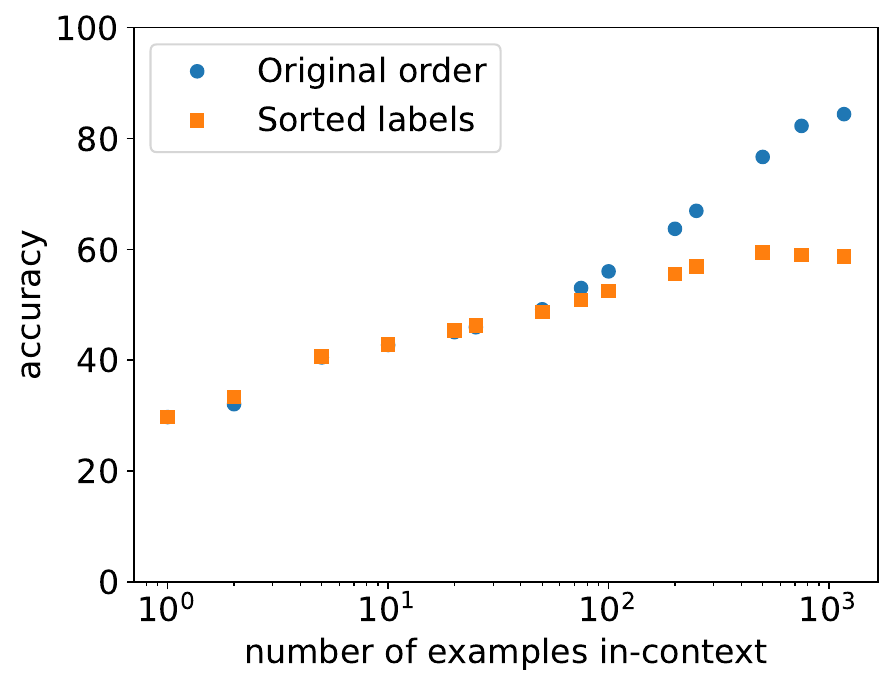} % Adjust width as needed
  \caption{By contrast, sorting examples by label has an increasingly negative impact on performance in longer context regimes. Results on \togetherllama with \clinic.}
  \label{fig:sorted-labels}
    \end{minipage}
\end{figure*}
We compare the properties of long-context ICL with the known properties of short-context ICL.\footnote{We consider using ICL as a testbed for properties of long-context models in Appendix~\ref{appendix:analysis-long-context}.} We primarily use classification, not generation, tasks to study these properties to avoid confounding effects from the difficulty of evaluating generated texts.

\paragraph{Is it best to use the entire context?} 
Prior work suggested that, for some simple tasks, providing additional input can \textit{reduce} performance \citep{levy2024task}. However, we observe monotonically increasing performance on nearly every dataset; in cases where the performance curve begins to flatten, small variation occurs, but no significantly lower performance occurs at higher example counts. While using the full context window is computationally costly and may be unnecessary to achieve high performance on some datasets, it is minimally not harmful to performance.

\paragraph{Sensitivity to example order}
Many models exhibit strong sensitivity to example order in-context \citep{lu-etal-2022-fantastically}. We examine this by measuring the percentage of predictions that change when the input is reordered (averaged over 3 shuffles). Figure~\ref{fig:reording-demonstrations} shows that, while some sensitivity to order persists, this effect weakens substantially with longer context. Across all datasets, the percent of labels flipped by shuffling in 1000-shot ICL is less than \textit{half} the percentage flipped in 10-shot ICL.

\paragraph{Label sorting}
We also consider an adversarial case for example ordering: we sort the examples so that examples with the same label appear together. At small numbers of examples, this has very little impact; if the average number of examples per class is low, label sorting is similar to a random sort. However, as the number of examples grows, label sorting begins to have a dramatic impact on performance. Figure~\ref{fig:sorted-labels} shows the performance of \togetherllama on \clinic with and without label sorting. As the number of examples in-context increases, the penalty for input sorting increases as well; at 1169-shot ICL, label sorting decreases accuracy by 25.7 percentage points. This suggests that contextualization of examples with \emph{different} labels is important to performance, and that this contextualization only occurs effectively over relatively short distances in the context window.

\begin{figure}[h]
    \centering
    \includegraphics[width=0.9\linewidth]{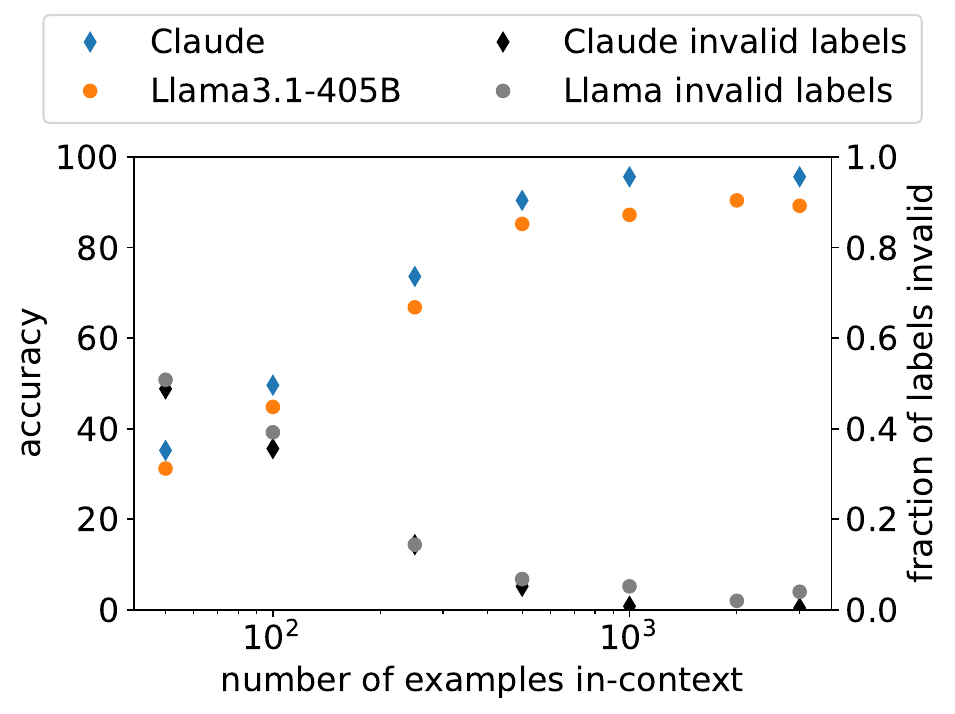}
    \caption{Performance of two frontier models on \clinic. Performance increases with the number of demonstrations at first, but saturates relatively early; the number of invalid labels produced continues to decline with increased demonstrations, even when accuracy plateaus.}
    \label{fig:frontier-models}
\end{figure}

\paragraph{Effectiveness for frontier models}
We focus on 7B/8B models because of the feasibility of in-depth analysis; however, it is also useful to consider the performance of the current strongest long-context models. We evaluate Claude Sonnet 3.5 \cite{claude35sonnet2024} and Llama 3.1 405B \cite{dubey2024llama3herdmodels}, using 50 to 3000 demonstrations from \clinic. Due to computational cost and (in the case of Claude) the limitations of API access, we do not apply constrained decoding in these runs and use only a single run over the test set instead of 10 runs at each example count. Figure~\ref{fig:frontier-models} shows these results. While performance increases past the typical fewshot range, at least some of this benefit comes from improved knowledge of the label space (i.e. less invalid labels generated) and performance on this task saturates quite quickly. Thus, while long-context ICL seems to be a promising direction even for frontier models, there may be diminishing returns on relatively simple tasks such as intent classification at even the 1,000-example scale.

\paragraph{Tasks where long context does not help} Concurrently to our work, \citet{li2024longcontext} identify a set of tasks where long context is not uniformly helpful. However, we observe that the tasks that show this trend either have near-0 performance at short demonstration lengths or also display an inverse performance trend on the short context scale (e.g. for TacRED \citep{zhang-etal-2017-position}, we observe that performance decreases from 1 to 10 total demonstrations; on Discovery \citep{sileo-etal-2019-mining}, we observe that performance is near-zero at all demonstration lengths and decreases from 5 to 10 total demonstrations). While these are important failure modes of language models, we restrict our analysis to tasks without these confounding issues. In \banking, the one dataset that our works share, both papers observe similar trends of improved performance with additional context. 

\section{Why does long-context ICL help?}
\label{sec:why}

The most notable differences between long-context and short-context ICL are the additional number of demonstrations and the average number of demonstrations that each demonstration is contextualized \textit{with respect to}-- that is, the same demonstration may have a different impact on the prediction if it is the only demonstration in-context versus if it is the 90th demonstration to appear in-context, because the demonstration will be better-contextualized.

We hypothesize that long-context ICL is primarily beneficial because of \textit{retrieval in-context}-- the idea that attention is used to select examples, rather than the model aggregating a complex decision boundary by better contextualizing across examples. This means that the primary benefit of long-context ICL is the number of demonstrations, \textit{not} the quality of contextualization that each individual demonstration receives. If this is the case, then there should exist some sparse attention pattern that severely restricts long-range attention \textit{between} demonstrations, but has limited or no impact on ICL performance.\footnote{Note that this is distinct from methods that overload the same positions with multiple embeddings in order to process longer contexts (e.g. \cite{Ratner2022ParallelCW}); here, we are not modifying any positional information, only restricting attention between demonstrations to a local context block.}  In all cases, the current test example can attend back to all demonstrations, in the same style as \citet{acharya2024starattentionefficientllm}. 

\textbf{Block-sparse attention patterns} 
We test several previously proposed sparse attention patterns and observe that two things are necessary to enable block attention: attending to an \emph{attention sink} \cite{xiao2024efficient} block that is always first in the context window; and attending to two prior local blocks \cite{guo2024blocksparse}. We ablate over these decisions in Appendix~\ref{appendix:block-attn-ablations}.

The pattern we identify, a small variation on Star Attention \cite{acharya2024starattentionefficientllm}, is visualized in Figure~\ref{fig:attn-patterns}. This attention pattern sparsifies attention between demonstrations substantially, removing almost all long-range connectivity between demonstrations in-context, without substantial impact on performance. This suggests that long-context ICL does not, in the encoding of the demonstration set, require long-range attention.

This pattern can also be used to disentangle two previously conflated factors: \textbf{$k$}, the total number of demonstrations that the test example can attend back to (e.g. $k$-shot prompting); and \textbf{$b$}, the number of examples per block in the blockwise attention pattern.\footnote{Note that each block attends to two blocks of local context ($2b$ examples) and a sink block ($b$ examples), so that the maximum number of demonstrations that the \textit{last} demonstration in the last block can attend to is actually $4b-1$.} block of examples that are contextualized together. We can fix the total number of examples in context, $k$, and vary $b$ by changing the attention mask; and we can fix $b$ with a custom attention mask and vary $k$ by adding additional blocks of examples to the context window. 

\begin{figure}

  \centering

\begin{subfigure}{0.45\linewidth}
  \centering
        \begin{tikzpicture}[scale=0.22]    
        \def\N{15}  % Number of tokens

        % Causal attention pattern
        \foreach \i in {1,...,\N} {
            \foreach \j in {1,...,\i} {
                \fill[Cyan!30] (\j, -\i) rectangle ++(1,-1);
            }
        }
        
        % Full attention for last token
        \foreach \j in {1,...,\N} {
            \fill[Cyan!30] (\j, -\N) rectangle ++(1,-1);
        }
        
        % Dark blue diagonal
        \foreach \i in {1,...,\N} {
            \fill[NavyBlue!60] (\i, -\i) rectangle ++(1,-1);
        }

        % Draw the grid
        \foreach \i in {1,...,\N} {
            \foreach \j in {1,...,\N} {
                \draw[gray!70] (\i, -\j) rectangle ++(1,-1);
            }
        }
                
            \end{tikzpicture}

    \caption{Causal attention}
    \label{fig:enter-label}
    \end{subfigure}
    \hspace{0.04\linewidth}
\begin{subfigure}{0.45\linewidth}
  \centering
        \begin{tikzpicture}[scale=0.22]    
        \def\N{15}  % Number of tokens
        \def\B{2}   % Block size

          % Sink attention pattern
        \foreach \i in {1,...,\N} {
            \fill[Cyan!30] (1, -\i) rectangle ++(1,-1);
        }

        % local blocks
        \foreach \i in {3,...,\N} {
            \fill[Cyan!30] (\i, -\i) rectangle ++(-1,-1);
        }

        \foreach \i in {4,...,\N} {
            \fill[Cyan!30] (\i, -\i) rectangle ++(-2,-1);
        }

        % Full attention for last block
        \foreach \j in {1,...,\N} {
            \fill[Cyan!30] (\j, -\N) rectangle ++(1,-1);
        }
    
    % Dark blue diagonal
        \foreach \i in {1,...,\N} {
            \fill[NavyBlue!60] (\i, -\i) rectangle ++(1,-1);
        }

        % Draw the grid
        \foreach \i in {1,...,\N} {
            \foreach \j in {1,...,\N} {
                \draw[gray!70] (\i, -\j) rectangle ++(1,-1);
            }
        }
        \end{tikzpicture}

    \caption{Blockwise attention}
    \end{subfigure}
    \caption{Causal attention versus the blockwise pattern we apply in the following experiments. Here, each square represents a \textit{block} of examples, and the last square represents the test example. This pattern represents attending to the first block (the attention sink) and two local blocks. }
    \label{fig:attn-patterns}

\end{figure}

\begin{figure}[h!]
  \centering
  \centering
  \includegraphics[width=0.9\linewidth]{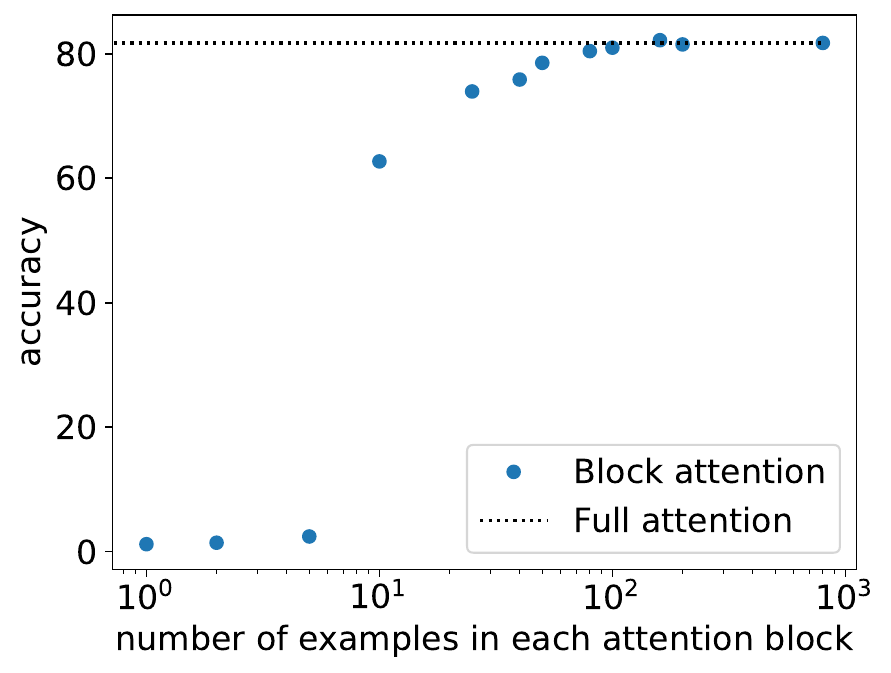}
\caption{Comparing block attention to full attention over the same set of examples with \togetherllama on \banking (i.e., fixing $k$ and varying $b$ along the x axis). Block attention approaches full attention (the black skyline) with relatively small block size.}
 \label{fig:banking-block}
\end{figure}

\textbf{Restricting contextualization on a fixed example set}
We fix a number of examples per-block. If the block size is equal to the number of examples in-context ($b=k$), this is equivalent to normal (full) attention; if the block size is $b=1$, each example can attend only to itself and its sink/local block.
Figure~\ref{fig:banking-block} shows results on \banking as a representative example. 
The performance of block attention quickly approaches the performance of full attention;  95\% of the performance of full attention is recovered by a block of $50$ examples in the case of \banking  (and in a similar range for all datasets studied). This suggests that the ICL performance is not strongly benefiting from encoding long-range (or even medium-range) dependencies across the demonstration set. However, some short-range dependencies between examples are clearly necessary for ICL performance, as using block sizes $b < 10$ results in near-zero performance and performance increases slightly with increased size from there. 

\begin{figure}[h!]
    \centering
  \includegraphics[width=0.9\linewidth]{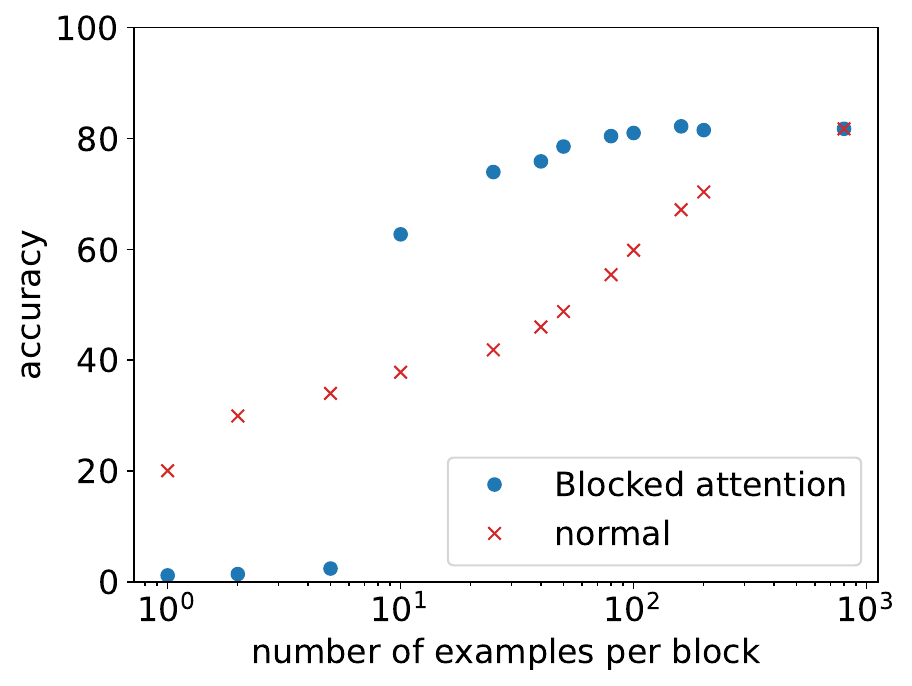 }

\caption{Comparing a single block to many blocks with a fixed block size using \togetherllama on \banking (i.e., fixing $b$ and varying $k$). When the block size is very small, poor contextualization appears to be the limiting factor on performance (so adding more examples with the same block size does not help); when the block size is larger, the total number of examples appears to be the limiting factor on performance (so adding more examples with the same block size is helpful).}
  \label{fig:banking-comp-k}
\vspace{-10pt}
\end{figure}

\textbf{Increasing number of examples with a fixed contextualization quality}
We fix the block size and compare attending to a single block ($k=b$) to attending to many blocks ($k>>b$) in Figure~\ref{fig:banking-comp-k}. This comparison measures the effect of adding additional examples without introducing substantial improvements to the contextualization of each individual example. 

When the context is extremely local (e.g. for banking, $b<10$ examples), attending across many locally encoded blocks is \emph{worse} than attending to a single block of examples. We hypothesize that this is due to inadequate contextualization of each example leading to less informative embeddings. 
However, after some minimal average quality of contextualization is achieved (in Figure~\ref{fig:banking-comp-k}, around $b=10$), adding more blocks of examples with the same local encoding dramatically increases performance. 
This supports the hypothesis that the primary performance improvement from long-context modeling is due to retrieving from more relevant examples in-context, rather than learning a better task boundary; this is supported as well by the retrieval results in Table~\ref{tab:results-main}, where retrieval performance at short contexts is near (though never exceeding) very-long-context ICL performance.

\section{Related Work}
\label{sec:related-work}
\textbf{Augmenting decoder-only models with long context} \quad Many methods for extending the context of language models have been introduced in the last few years. 
One series of work has focused on positional embedding extrapolation strategies \citep{peng2023yarn, rozière2024code, chen2023extending, liu2023ring, zhu2024pose, xiao2024efficient, han2024lminfinite}. 
When extrapolating past pretraining length, models also generally benefit from additional finetuning on long-context data \citep{xiong2023effective}. 
Other methods include adding retrieval-based attention \citep{unlimiformer, focused_transformer, yen2024longcontext} or hierarchical merging of information \citep{song2024hierarchical, yu2023megabyte}.
The two long-context Llama variants we consider in this work are both examples of finetuning for length extrapolation.

Separately, methods for longer context for ICL have also been proposed. Parallel context windows \citep{Ratner2022ParallelCW} and structured prompting \citep{Hao2022StructuredPS} propose methods of re-using the same positional embeddings multiple times to encode more demonstrations; this is quite effective for small numbers of overlaps, albeit with diminishing returns as the number of overlapping windows increases. \citet{li2023incontext} propose a new efficient attention mechanism and motivate its use using long-context ICL on a model tuned with many-shot instruction finetuning. \citet{cho2023promptaugmented} propose a hybrid of ICL and linear prompting which improves beyond few-shot ICL performance.  

Several works have also critiqued the efficacy of long context models. 
\citet{lost_in_the_middle} demonstrate that some long-context models fail to effectively use the middle of the context window; the models we use were released after this work and have generally high scores for middle-of-context retrieval. 
\citet{li2023how} suggest that some long-context models are only effective at utilizing inputs that are shorter than their context window's intended supported length; we do not observe this effect strongly, though we do see saturating performance slightly before the maximum number of examples in-context for some models.
\citet{li2023loogle} show that many models fail at tasks that require reasoning over long dependency lengths; this is unlikely to be an issue in our setting.

\textbf{Properties of in-context learning} \quad
\citet{milios-etal-2023-context} study ICL for many-class classification with models up to 4k context length and find that, when retrieving demonstrations, 7b models show early performance saturation on many tasks. 
Our results suggest that this failure to use longer context effectively is not an inherent property of 7b models, but instead a type of shallow heuristic used by some particular models when the demonstrations are of sufficiently high quality.  

\citet{xu2023understanding} study the impacts of ground-truth label, input distribution, and explanations on ICL performance; 
\citet{bolucu-etal-2023-impact} study the impact of example selection in a specific domain.
\citet{lin2024dual} argue that ICL occurs in two modes: learning tasks and retrieving tasks, and that retrieval of similar-but-not-quite-correct tasks can explain ``early ascent'' behaviors where ICL performance peaks once in a fewshot regime and then performance improves again with a much higher number of examples. Similarly, \citet{pan-etal-2023-context} argue for a distinction between task recognition and task learning, and suggest that task learning continues to benefit from additional examples at scale.
\citet{vonoswald2023transformers} suggest in-context learning can be viewed as gradient descent, although \citet{deutch2024incontext} argue against this interpretation. 
 \citet{hendel-etal-2023-context} view ICL as compressing the demonstrations into a ``task vector'' that maps from inputs to outputs. 
 
Concurrently to our work, \citet{agarwal2024manyshot} study many-shot prompting of Gemini 1.5 and show improvements from the fewshot setting across both classification and generation tasks. Our work differs in its evaluation of multiple open-source models, our
comparison to finetuning the same base model, and our use of ICL as a testbed for analysis of long context behaviors.

\textbf{Comparing in-context learning and finetuning}\quad \citet{min-etal-2022-metaicl} show that models trained on fewshot learning can generalize to perform fewshot learning on new tasks; in some cases, this can outperform finetuning directly on the new task.
\citet{mosbach-etal-2023-shot} compare finetuning to ICL more directly; they find that finetuning generally outperforms ICL with the same number of examples both in-domain and out-of-domain, when comparing 16-example ICL to finetuning on the same 16 examples. Their setting differs from ours in their choice of model (OPT) and the amount of data considered (16 for ICL, 16 or 128 for finetuning).
\citet{liu2022fewshot} find that PEFT generally outperforms ICL in their setting, where they finetune an encoder-decoder model with a language modeling objective using their T-few method and 20-70 samples. 
\citet{asai2023buffet} compare finetuning and ICL for mT5 on cross-lingual transfer and find that ICL outperforms finetuning in some, but not all, of the tasks studied.
To the best of our knowledge, no prior work has considered the relative performance of finetuning and ICL in the many-shot regime with hundreds or thousands of examples in-context.

\section{Conclusion}
We have demonstrated that ICL with large demonstration sets can be surprisingly effective, and shed light on a few surprising properties in its behavior. Namely, long-context ICL exhibits a reduced dependence on example selection, relatively stable performance with respect to example order, and performance often approaching or exceeding parameter-efficient finetuning on the same data, all properties that make this an appealing option for a variety of tasks. 
We have also shown that long-context ICL's effectiveness is largely due to retrieval from the long context during prediction, rather than cross-attention within the large demonstration set during encoding. 

Our work also highlights that our understanding of ICL remains incomplete. Though much work has studied the potential mechanisms behind ICL, these works have largely focused on simple tasks with small ($<10$ examples) demonstration sets; as our work demonstrates that ICL's properties shift in the long context regime, more work is necessary to validate hypotheses about ICL at larger scales. 

While prior work has focused on two strategies for performing inference on a new task-- either finetuning on task-specific data or selecting a subset of that data to use in-context-- our results points to a potential third paradigm: adapting the \textit{model} to fit as much of that data in-context as possible, caching and reusing the encoding of the long demonstration set. While finetuning with full datasets is still a powerful option if the data vastly exceeds the context length, our results suggest that long-context ICL is an effective alternative. ICL trades finetuning-time cost for increased inference-time compute, and increasing the amount of inference-time compute by using more examples in-context is an effective strategy to improve performance.
As the effectiveness and efficiency of long context models continue to increase, we believe long-context ICL will be a powerful tool for many tasks.

\section{Limitations}
Our work focuses on open-source models (and predominately on the Llama-2 family); more work is necessary to establish whether this trend holds across model families, although we are encouraged by the strong results \citet{agarwal2024manyshot} observed for many-shot ICL on Gemini 1.5. Additionally, we do not consider other non-LoRA PEFT methodologies; it is possible that some of these methodologies may outperform in-context learning. Finally, we focus primarily on classification tasks, and care should be taken in generalizing to new tasks. 

\section{Broader impacts}
Any work that studies general capabilities of language models can be used for both positive and negative applications downstream. In-context learning work is perhaps particularly vulnerable to this dual use, in part because it is a relatively accessible method, requiring far less compute than finetuning or pretraining models.

Independent of our work, \citet{anil2024manyshot} observed that many-shot prompting can be used to jailbreak some models. We do not study or suggest the use of long-context ICL for jailbreaking in our work, but this is a potentially harmful use case of this paradigm. Outside of this risk, we do not foresee any additional harms introduced by long-context ICL methods.

We hope that our work opens up additional possibilities for people who are compute-constrained to customize models to their use cases or tasks. We also aim to carefully explore the boundary between where finetuning and long-context ICL is appropriate, to enable practitioners to make informed choices on when to apply each method.

\subsection*{Acknowledgments}
We would like to thank  Vijay Viswanathan, Sewon Min, Akari Asai, Xiang Yue, and Simran Khanuja for useful discussions about this work.

This work was partially supported by The Yandex Initiative for Machine Learning, the Len Blavatnik and the Blavatnik Family foundation, and the European Research Council (ERC) under the European Union Horizons 2020 research and innovation programme (grant ERC DELPHI 802800). AB was supported by a grant from the National Science Foundation Graduate Research Fellowship Program under Grant No. DGE2140739. MI also acknowledges the support of the Israeli Council of Higher Education. 
Any opinions, findings, and conclusions or recommendations expressed in this material are those of the author(s) and do not necessarily reflect the views of the sponsors.

\bibliography{arxiv_version}
\bibliographystyle{colm2024_conference}

\newpage
\appendix
\onecolumn
\section{Saturation}
\label{appendix:saturation}
One metric we are interested in is the point where the model performance \emph{saturates}, which we define informally as the point where adding more examples is unlikely to meaningfully improve performance. More formally, we define the saturation point as the smallest number of examples tested such that performance reaches 95\% of the model's maximum performance. 
\paragraph{Saturation points vary by dataset.}
We define saturation as the first point at which performance reaches 95\% of the model's maximum performance on that dataset. Table~\ref{saturation-table} shows the number of examples at saturation and the maximum number of examples that fit in the context window for each model. For datasets with larger label spaces, saturation generally occurs later; \banking and \clinic   do not saturate within the context window of Llama2 (4096 tokens, which represents between 100-162 in-context examples for these datasets). In the longer-context regime, saturation points generally occur slightly later on \longllama, but in both models occur before the model's maximum context length. 

This suggests two things. First, given a fixed model, it is often not necessary to use the full context length to extract high performance from that model. Second, current models do not make use of the full potential of ICL; models often saturate in performance before the maximum number of examples, despite longer-context versions revealing that further performance improvements are possible.

\paragraph{The number of classes has some impact on saturation point-- but is not fully explanatory.}
Our results show datasets with more classes benefit from more demonstrations in-context, on average, before saturation. This is to be expected, as the expected number of demonstrations necessary before seeing the correct label increases with the number of total label classes. To test if this is an intrinsic property of these datasets, or truly linked only to the number of label classes, we construct subsets of two high-label-space datasets, \banking and \clinic, by randomly selecting half of the labels to exclude from the dataset. These subsets remain in the same domain, but with a smaller label space; if saturation is tied to the number of examples, then this should move the saturation point. Note that this is distinct from \textit{combining} labels (e.g. TREC vs TREC-FINE), as combining finegrained labels into general labels makes the classification task simpler. It's still possible that the subset chosen is a simpler task (e.g. by removing one of a pair of frequently confused labels); to moderate the effect of this change, we average results over 3 randomly chosen subsets. 

Table~\ref{tab:half-label-space} compares the saturation point of the full- and half-label-space runs. For the datasets with the most number of labels, halving the number of labels also reduces the amount of examples that are useful before saturation, albeit not by half. However, the trend is less clear for the tasks with fewer labels; in some cases, reducing the label space actually \textit{increases} the number of  demonstrations before saturation. While the size of the label space clearly has some impact on the saturation point, more investigation is necessary to identify other factors impacting this behavior.

\begin{table*}[h]
\begin{center}
\begin{tabular}{lllll}
\toprule
\multicolumn{1}{c}{\bf Dataset}  &\multicolumn{1}{c}{\bf Llama2}  &\multicolumn{1}{c}{\bf Llama2-32k}  &\multicolumn{1}{c}{\bf Llama2-80k}  & {\bf Mistral} \\ 
\midrule 

\trec & 20 (140) & 100 (1129) & 75 (2000) & 50 (1129)  \\
\trecfine & 75 (131) & 250 (1056) & 500 (2000) & 500 (1091)  \\
\nlu & 100 (162) & 500 (1309) & 500 (2000) & 250 (1309)  \\
\banking & - (100) & 500 (838) & 750 (1750) & 500 (860)  \\
\clinic & - (145) & 750 (1169) & 1000 (2000) & 750 (1212)  \\
\bottomrule
\end{tabular}
\end{center}
\caption{We measure the saturation point as the point at which the model reaches 95\% of its maximum accuracy on the dataset; ``-'' in a column indicates that the maximum performance is achieved by using the full context window. The number in parenthesis represents the maximum number of examples that fit in the context window. As the label space of the dataset increases (from top to bottom row), so does the number of examples that can be used before saturation.}
\label{saturation-table}

\end{table*}

\begin{table*}
\begin{center}
\begin{tabular}{lllll}
\toprule
\multicolumn{1}{c}{\bf Dataset}  &\multicolumn{1}{c}{\bf Llama2}  &\multicolumn{1}{c}{\bf Llama2-32k}  &\multicolumn{1}{c}{\bf Llama2-80k}  & {\bf Mistral} \\ 
\midrule 

\trec & 14.29 / 24.69 & 8.86 / 1.77 & 3.75 / 1.71 & 4.43 / 6.0  \\
\trecfine & 57.25 / 80.71 & 23.67 / 27.9 & 25.0 / 32.38 & 45.83 / 33.33  \\
\nlu & 61.73 / 80.71 & 38.2 / 17.21 & 25.0 / 26.67 & 19.1 / 36.67  \\
\banking & 100.0 / 88.0 & 59.67 / 37.91 & 42.86 / 28.57 & 58.14 / 35.56  \\
\clinic & 100.0 / 64.04 & 64.16 / 35.14 & 50.0 / 22.86 & 61.88 / 56.67  \\
\bottomrule
\end{tabular}
\end{center}
\caption{We compare the saturation point between the full-label-space (left) and half-label-space (right) for each model+dataset pair. Here we represent the label space as a percentage of the full context window.}
\label{tab:half-label-space}
\end{table*}

\newpage
\section{Using ICL as a testbed for long-context model properties}
\label{appendix:analysis-long-context}

In this section, we use in-context learning as a testbed to examine several properties of long-context models. 

\begin{figure}[h]
  \centering

\begin{subfigure}{.49\textwidth}
  \centering
  \includegraphics[width=\linewidth]{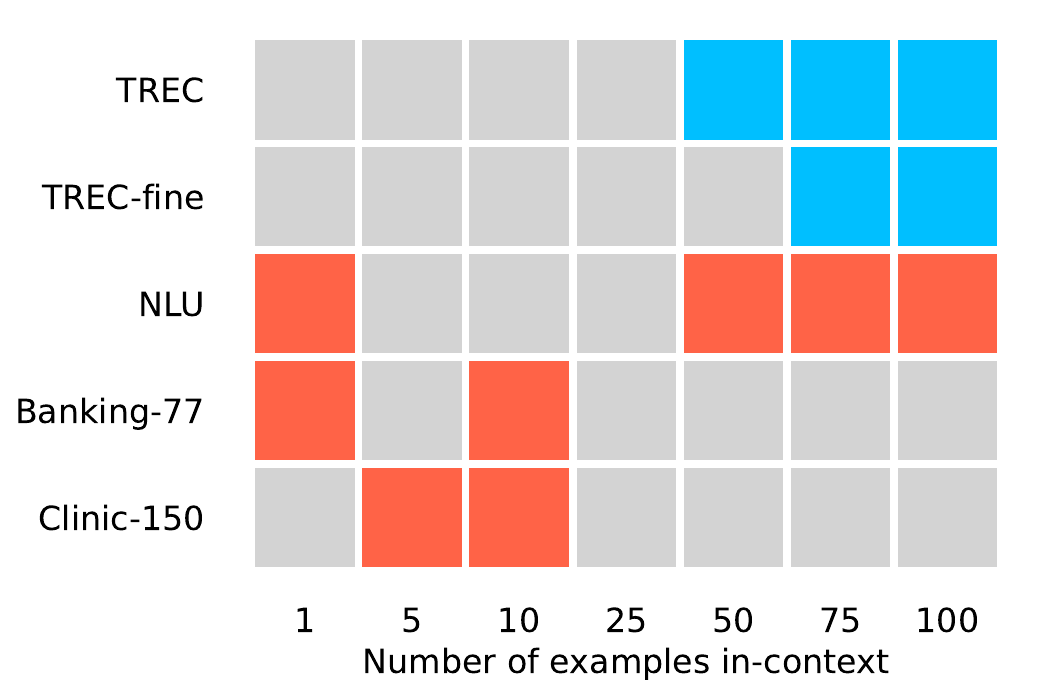}
  \caption{\togetherllama}
  \label{fig:32k-shortcontext}
\end{subfigure}
\begin{subfigure}{.49\textwidth}
  \centering
  \includegraphics[width=\linewidth]{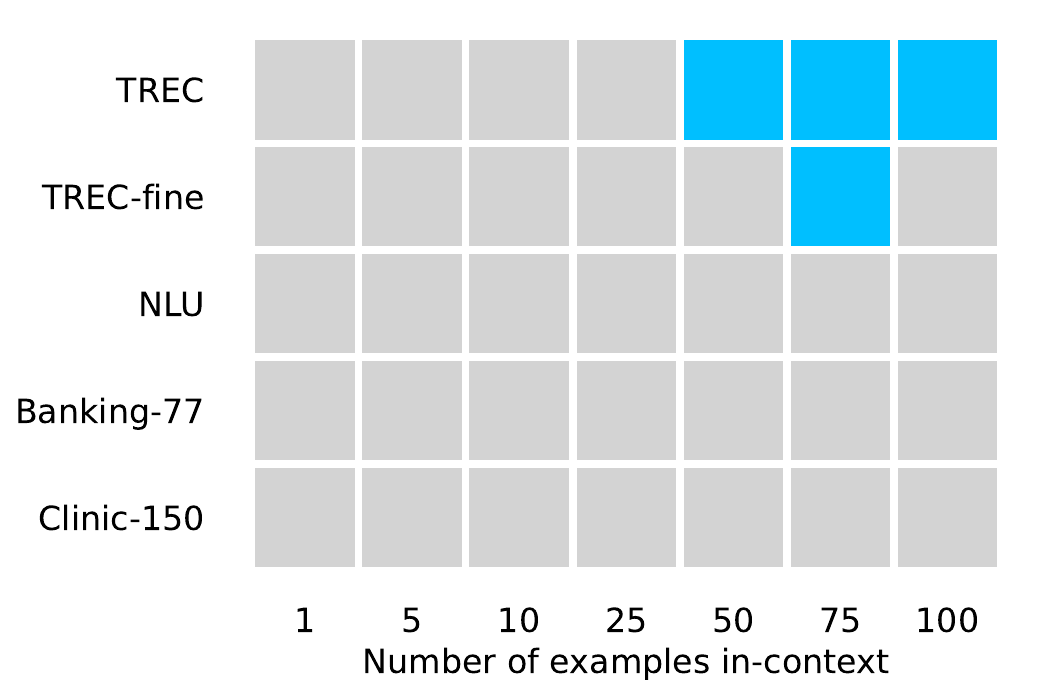}
  \caption{\longllama}
  \label{fig:80k-shortcontext}
\end{subfigure} \\
\caption{Short-context behavior of long-context models. Each model's performance is compared to the performance of the base model it was finetuned from, on the same amount of data. \textbf{\textcolor{red}{Red}} represents significantly worse performance; \textbf{\textcolor{blue}{blue}} represents significantly better performance ($p < 0.05$).}.
\label{fig:short-context}
\end{figure}

\paragraph{How do long-context models perform in the short-context regime?} \togetherllama and \longllama are finetuned variants of Llama2-7b, adapted for longer contexts. We evaluate how these models perform relative to the base model in short-context tasks (e.g. ICL using less than 4096 tokens of demonstrations) by testing whether the difference in performance is statistically significant (2-sided t-test, $p < 0.05$). Performance is generally similar, with some areas of slight improvement from the base model; Figure~\ref{fig:short-context} shows full results. We observe degradation in performance in some settings for \togetherllama, highlighting the importance of testing for behavior regression when finetuning for additional capabilities. 

\paragraph{Input utilization}

We analyze the performance of all methods on a naturalistic needle-in-the-haystack \cite{ivgi-etal-2023-efficient, lost_in_the_middle} style test.
If the model is effectively using the context, then it should be able to exactly recover the label for any example it has seen in-context. Note that, while a model trained on some set of data is not uniformly capable of exact copying from that training data \citep{biderman2023emergent}, in nearly all of our finetuning runs, the model fits the training data with 100\% accuracy. 

We examine this behavior by selecting the same set of examples to use in-context and in evaluation; all models should then be able to achieve 100\% accuracy. Table~\ref{tab:input-copying} shows the results; while all models achieve very high accuracy on the copied data, no model is able to uniformly copy correctly from the input. Surprisingly, performance improves slightly with additional demonstrations for most models, possibly due to additional specification of the task. 
\begin{table*}
\centering
\begin{tabular}{lllllllll}
\toprule
\textbf{Number of examples} & 1     & 5    & 10   & 25   & 50   & 100   & 200  & 250  \\ \midrule
Llama2              & \textbf{100.0} & 93.0 & 96.5 & 97.0 & 98.6 & 97.95 &    -  &  -    \\
\togetherllama                & 80.0  & 95.0 & 97.0 & 96.6 & 98.4 & 98.5  & 98.5 & \textbf{98.9} \\
\longllama                & 90.0  & 94.0 & 95.0 & 98.2 & \textbf{98.5} & 98.3  & 98.0 & 97.9
\end{tabular}
\caption{Copying behavior given the test examples in the context window. Results are averaged over \banking and \clinic; bold indicates the best performance for that model.}
\label{tab:input-copying}
\end{table*}

\newpage
\section{Full ICL results across datasets}
\label{appendix:all-figs}
For space, we show 1-2 representative datasets for each point of analysis in the paper. In this appendix, we present results across all classification datasets.
\newpage
\subsection{Random selection ICL across all models}
\begin{figure}[h!]
\begin{subfigure}{.5\textwidth}
  \centering
  \includegraphics[width=\linewidth]{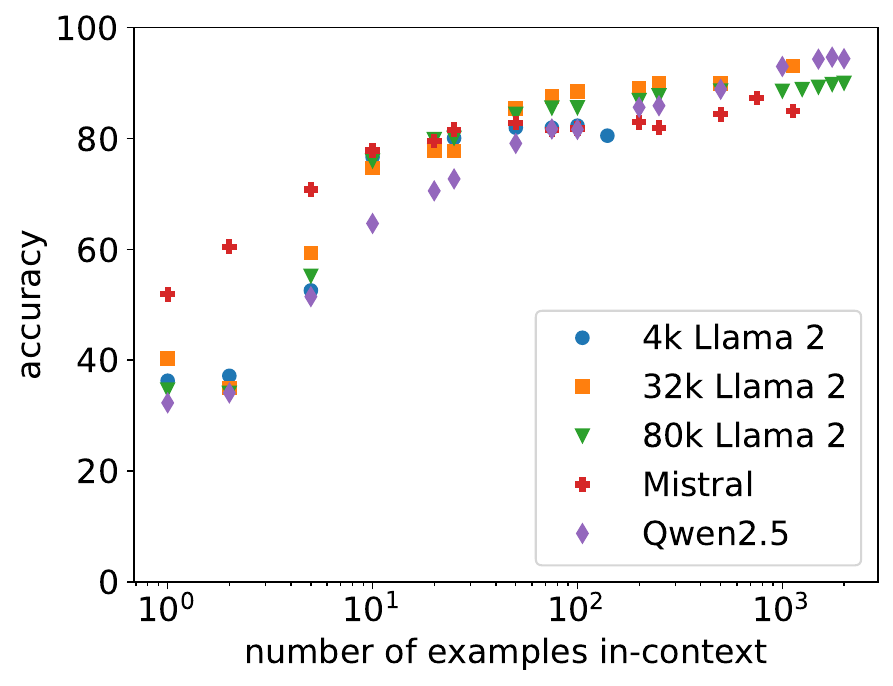}
  \caption{\trec}
  \label{fig:random-trec}
\end{subfigure}%
\begin{subfigure}{.5\textwidth}
  \centering
  \includegraphics[width=\linewidth]{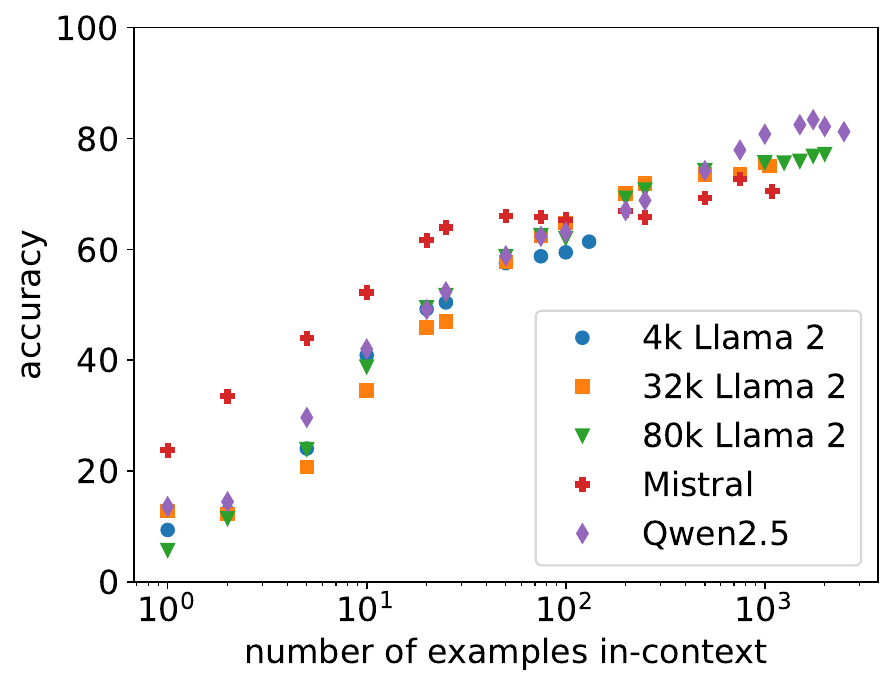}
  \caption{\trecfine}
  \label{fig:random-trecfine}
\end{subfigure} \\
\begin{subfigure}{.5\textwidth}
  \centering
  \includegraphics[width=\linewidth]{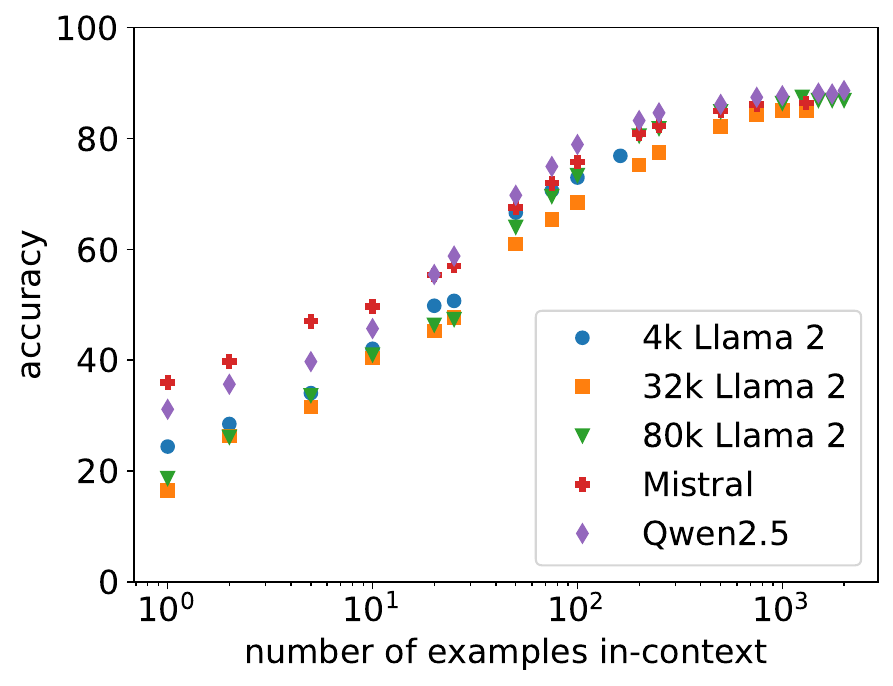}
  \caption{\nlu}
  \label{fig:random-nlu}
\end{subfigure}%
\begin{subfigure}{.5\textwidth}
  \centering
  \includegraphics[width=\linewidth]{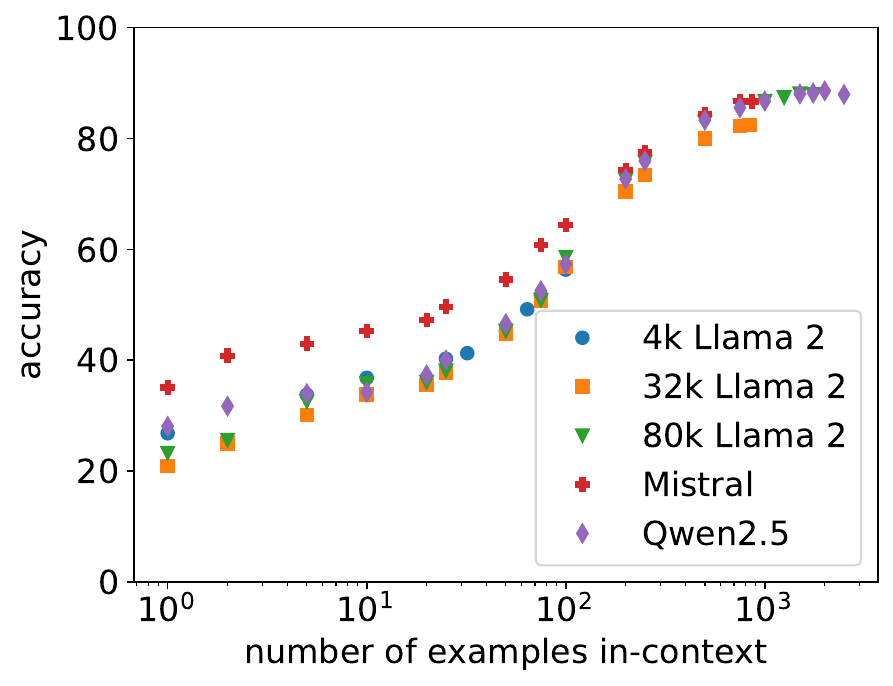}
  \caption{\banking}
  \label{fig:random-banking}
\end{subfigure}
\begin{subfigure}{.5\textwidth}
  \centering
  \includegraphics[width=\linewidth]{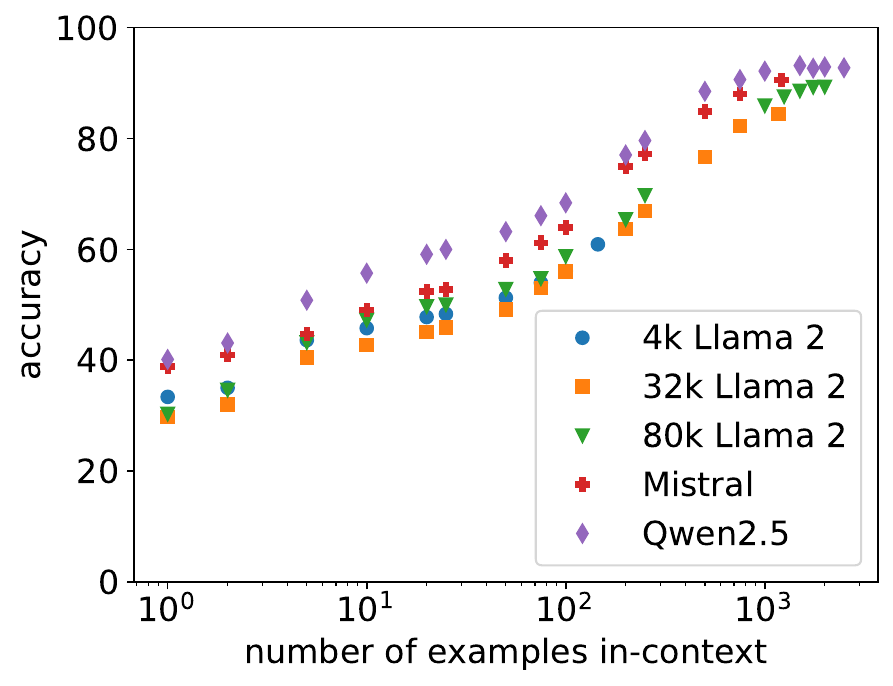}
  \caption{\clinic}
  \label{fig:random-clinic}
\end{subfigure}%
\caption{Performance of random-selection ICL across all models for each classification dataset. Performance continues to increase with additional examples in-context.}
\label{fig:all-models-random-ICL}
\end{figure} 

\newpage
\subsection{Retrieval ICL across all models}
\begin{figure}[h!]
\begin{subfigure}{.5\textwidth}
  \centering
  \includegraphics[width=\linewidth]{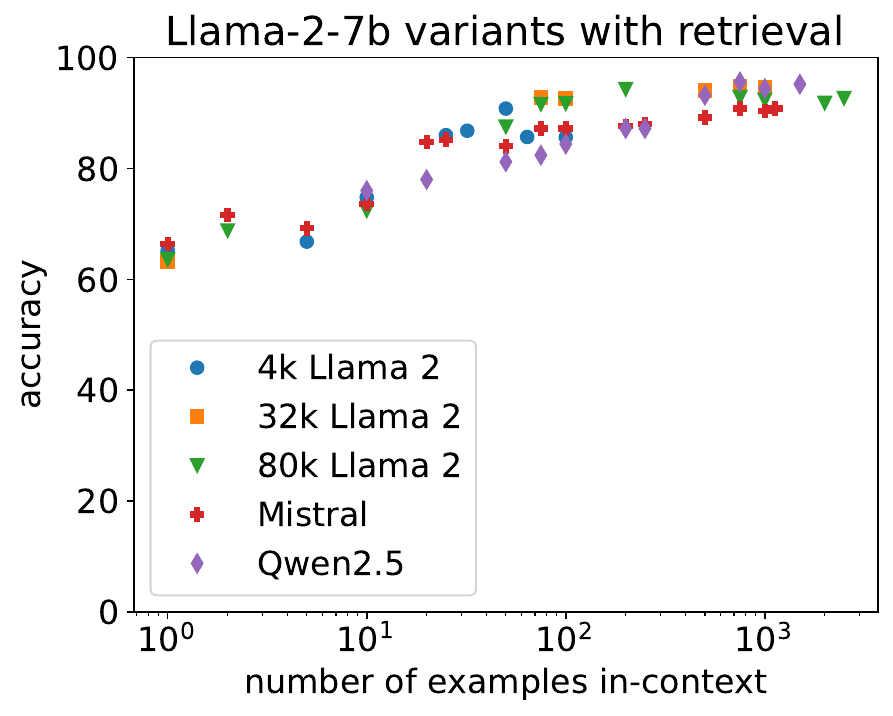}
  \caption{\trec}
  \label{fig:retrieval-trec}
\end{subfigure}%
\begin{subfigure}{.5\textwidth}
  \centering
  \includegraphics[width=\linewidth]{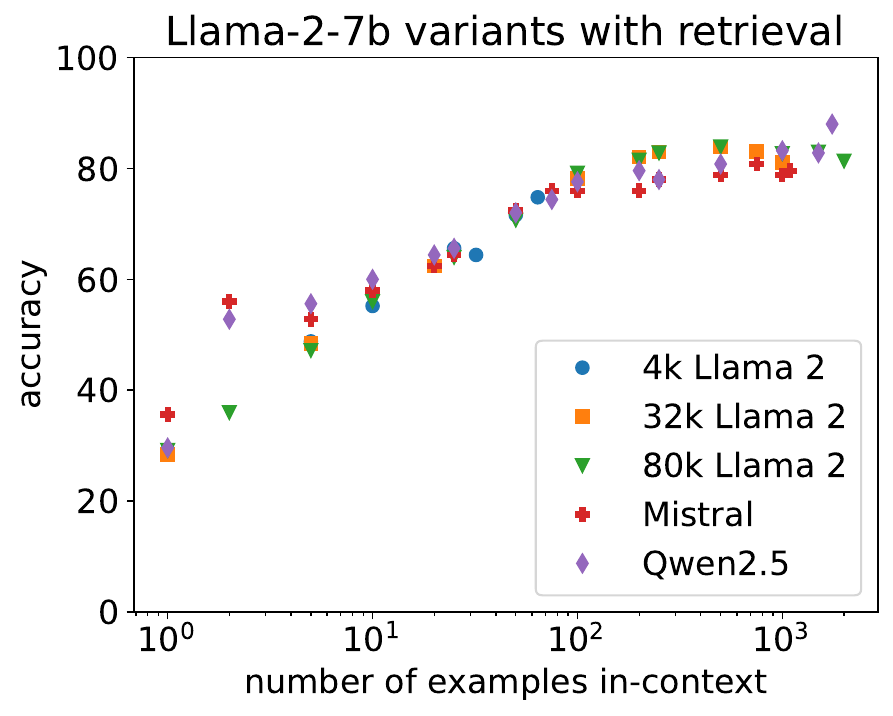}
  \caption{\trecfine}
  \label{fig:retrieval-trecfine}
\end{subfigure} \\
\begin{subfigure}{.5\textwidth}
  \centering
  \includegraphics[width=\linewidth]{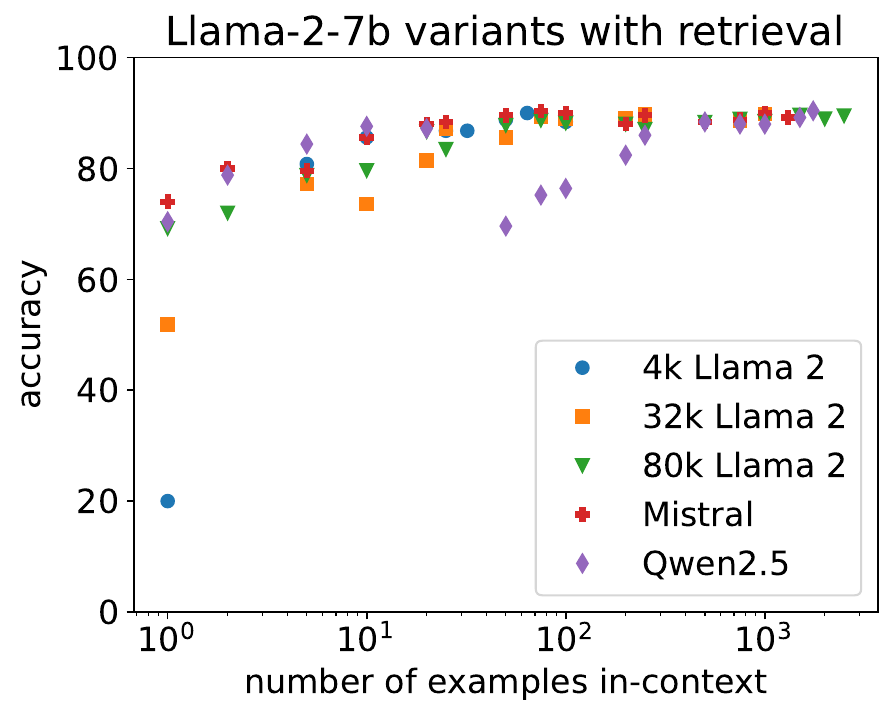}
  \caption{\nlu}
  \label{fig:retrieval-nlu}
\end{subfigure}%
\begin{subfigure}{.5\textwidth}
  \centering
  \includegraphics[width=\linewidth]{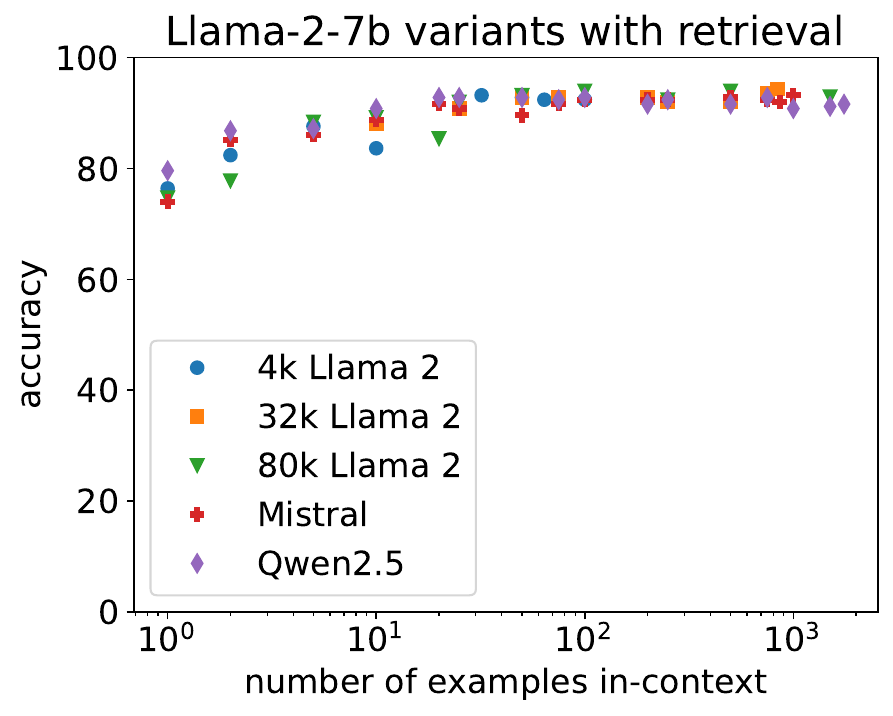}
  \caption{\banking}
  \label{fig:retrieval-banking}
\end{subfigure}
\begin{subfigure}{.5\textwidth}
  \centering
  \includegraphics[width=\linewidth]{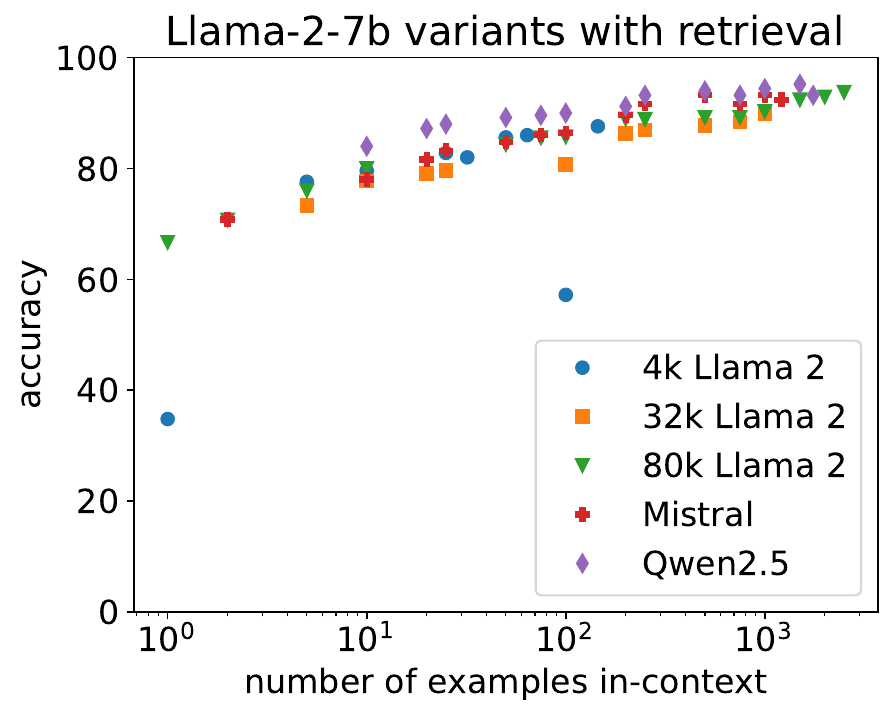}
  \caption{\clinic}
  \label{fig:retrieval-clinic}
\end{subfigure}%
\caption{Performance of retrieval-based ICL across all models for each classification dataset. Short-context performance here is higher than for random-selection, but performance continues to improve with more examples until a saturation point, where performance flattens out.}
\label{fig:all-models-retrieval-ICL}
\end{figure}

\newpage
\subsection{Comparing retrieval, random selection, and finetuning}
\begin{figure}[h!]
\begin{subfigure}{.5\linewidth}
  \centering
  \includegraphics[width=\linewidth]{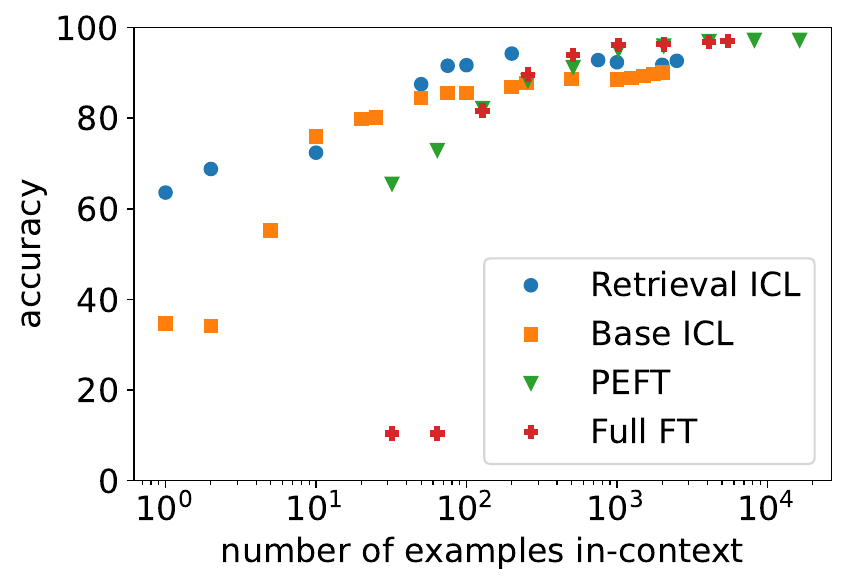}
  \caption{\trec}
  \label{fig:trec-all-comp}
\end{subfigure}%
\begin{subfigure}{.5\linewidth}
  \centering
  \includegraphics[width=\linewidth]{figures/compall/trecfine_all.pdf}
  \caption{\trecfine}
  \label{fig:trecfine-all-comp}
\end{subfigure} \\
\begin{subfigure}{.5\linewidth}
  \centering
  \includegraphics[width=\linewidth]{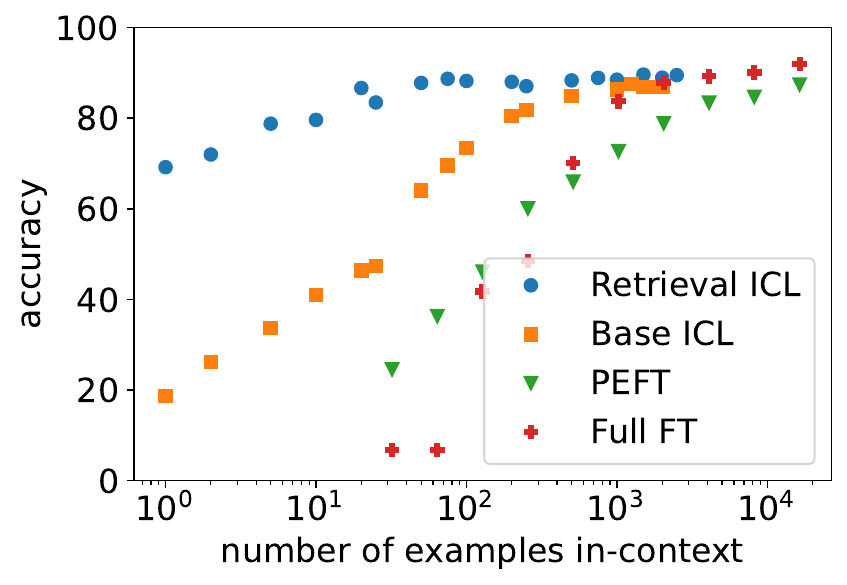}
  \caption{\nlu}
  \label{fig:nlu-all-comp}
\end{subfigure}
\begin{subfigure}{.5\linewidth}
  \centering
  \includegraphics[width=\linewidth]{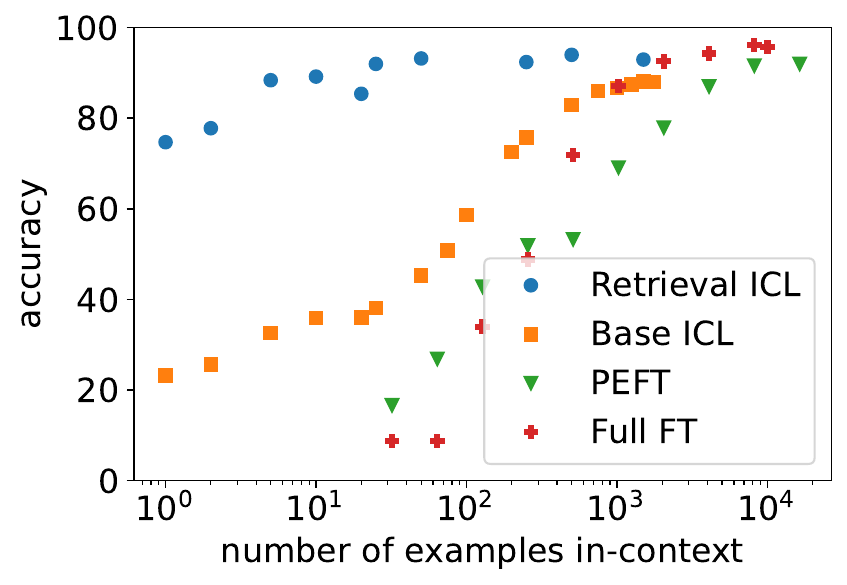}
  \caption{\banking}
  \label{fig:banking-all-comp} 
\end{subfigure} \\
\begin{subfigure}{.5\linewidth}
  \centering
  \includegraphics[width=\linewidth]{figures/compall/clinic_all.pdf}
  \caption{\clinic}
  \label{fig:clinic-all-comp}
\end{subfigure} 
\caption{Performance of retrieval-based ICL, random-selection ICL, and finetuning across 5 datasets. At small example counts, ICL outperforms finetuning; when several thousand examples are used, finetuning outperforms ICL in some datasets.}
\label{fig:all-datasets-finetuning-comp}
\end{figure}
\newpage

\section{Constrained decoding details}
\label{appendix:unconstrained}

\begin{figure}[h!]
\begin{subfigure}{.5\linewidth}
  \centering
  \includegraphics[width=\linewidth]{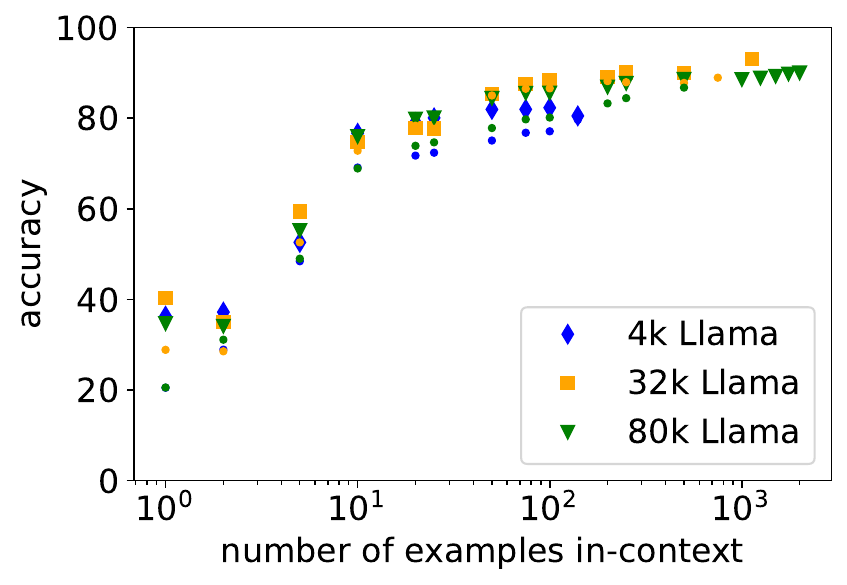}
  \caption{\trec}
  \label{fig:trec-unconstrained-comp}
\end{subfigure}%
\begin{subfigure}{.5\linewidth}
  \centering
  \includegraphics[width=\linewidth]{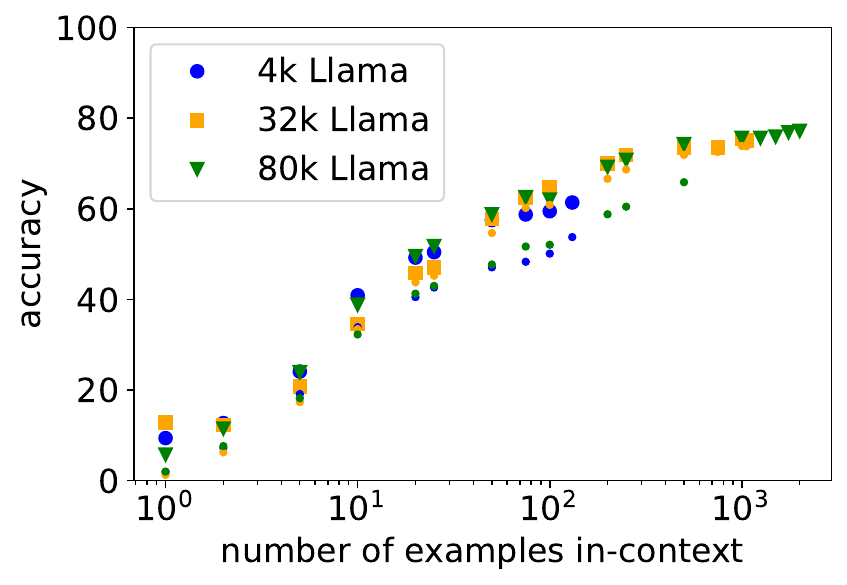}
  \caption{\trecfine}
  \label{fig:trecfine-unconstrained-comp}
\end{subfigure} \\
\begin{subfigure}{.5\linewidth}
  \centering
  \includegraphics[width=\linewidth]{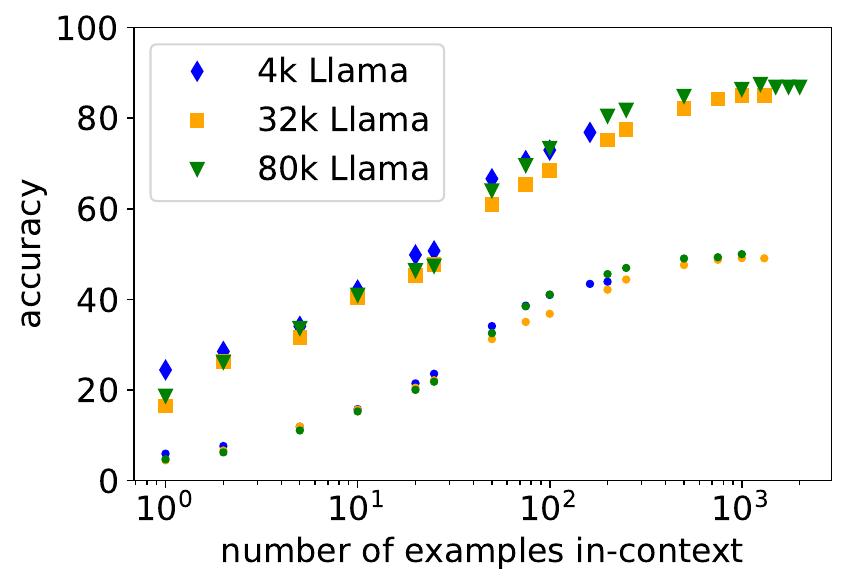}
  \caption{\nlu}
  \label{fig:nlu-unconstrained-comp}
\end{subfigure}
\begin{subfigure}{.5\linewidth}
  \centering
  \includegraphics[width=\linewidth]{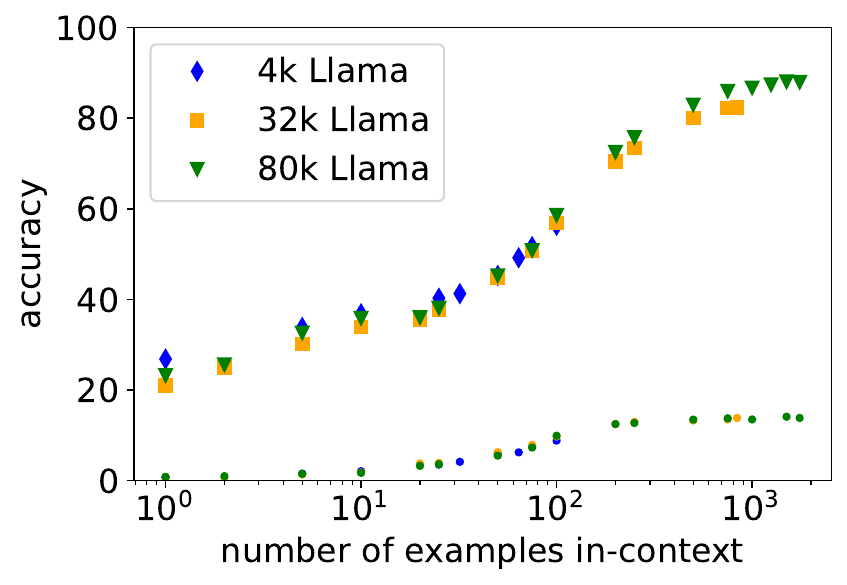}
  \caption{\banking}
  \label{fig:banking-unconstrained-comp} 
\end{subfigure} \\
\begin{subfigure}{.5\linewidth}
  \centering
  \includegraphics[width=\linewidth]{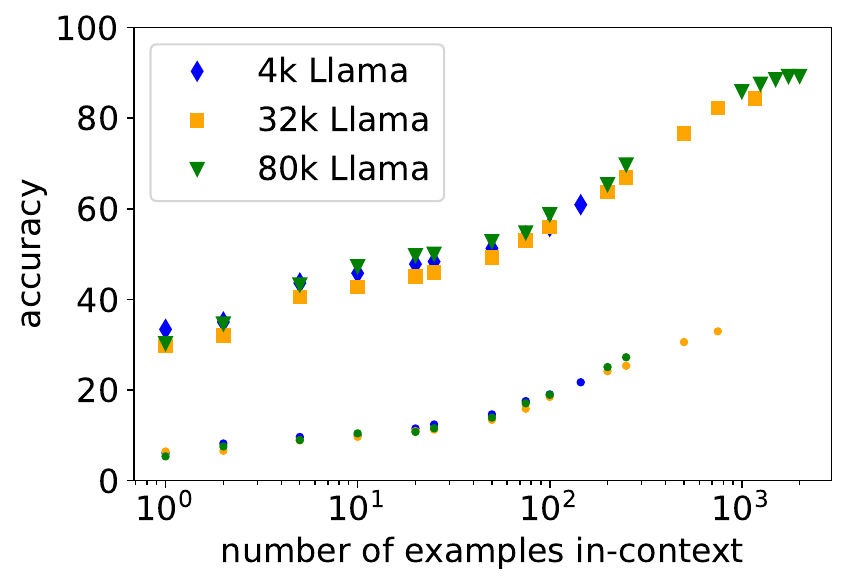}
  \caption{\clinic}
  \label{fig:clinic-unconstrained-comp}
\end{subfigure} 
\caption{Comparison of constrained decoding (diamond, square, and triangle datapoints on the graph) with the unconstrained decoding variants (dot-like datapoints of the same colors).}
\label{fig:all-datasets-unconstrained-comp}
\end{figure}

To perform constrained decoding, following \citet{Ratner2022ParallelCW}, we add a large constant to all logits at each decoding step that could result in a valid label being generated. This strategy is generally not possible for API-access models, or when the label space is not fully known in advance. To determine whether the same trends hold without the use of constrained decoding, we evaluated the Llama-2 family models with and without constrained decoding. Figure~\ref{fig:all-datasets-unconstrained-comp} shows the comparison; while performance is lower, especially in the higher-label-space datasets, the general trends hold. In Banking77, it appears that performance begins to saturate slightly earlier with unconstrained decoding; we hypothesize that this may be due to more specialized language used in banking domain labels.

\section{Finetuning}
\label{appendix:finetuning}
We performed finetuning with HuggingFace transformers \citep{wolf2020transformers}.
To perform parameter efficient finetuning (PEFT), we used the peft \citep{peft} package (version 0.9.0). 
We finetune the model for 30 epochs, evaluating it every epoch on the test set, and ultimately choosing the checkpoint with the highest test accuracy. We note that using the test set to perform model selection presents an unfair advantage to finetuning (compared to ICL) and may not be truly indicative of the generalization error. However, doing so provides the advantage of being both comparable to ICL in terms of the data used, as well as giving an upper bound on the true generalization accuracy of the finetuned model, further emphasizing any observed efficacy gap between it and ICL.

\begin{figure}
  \centering
  \includegraphics[width=\linewidth]{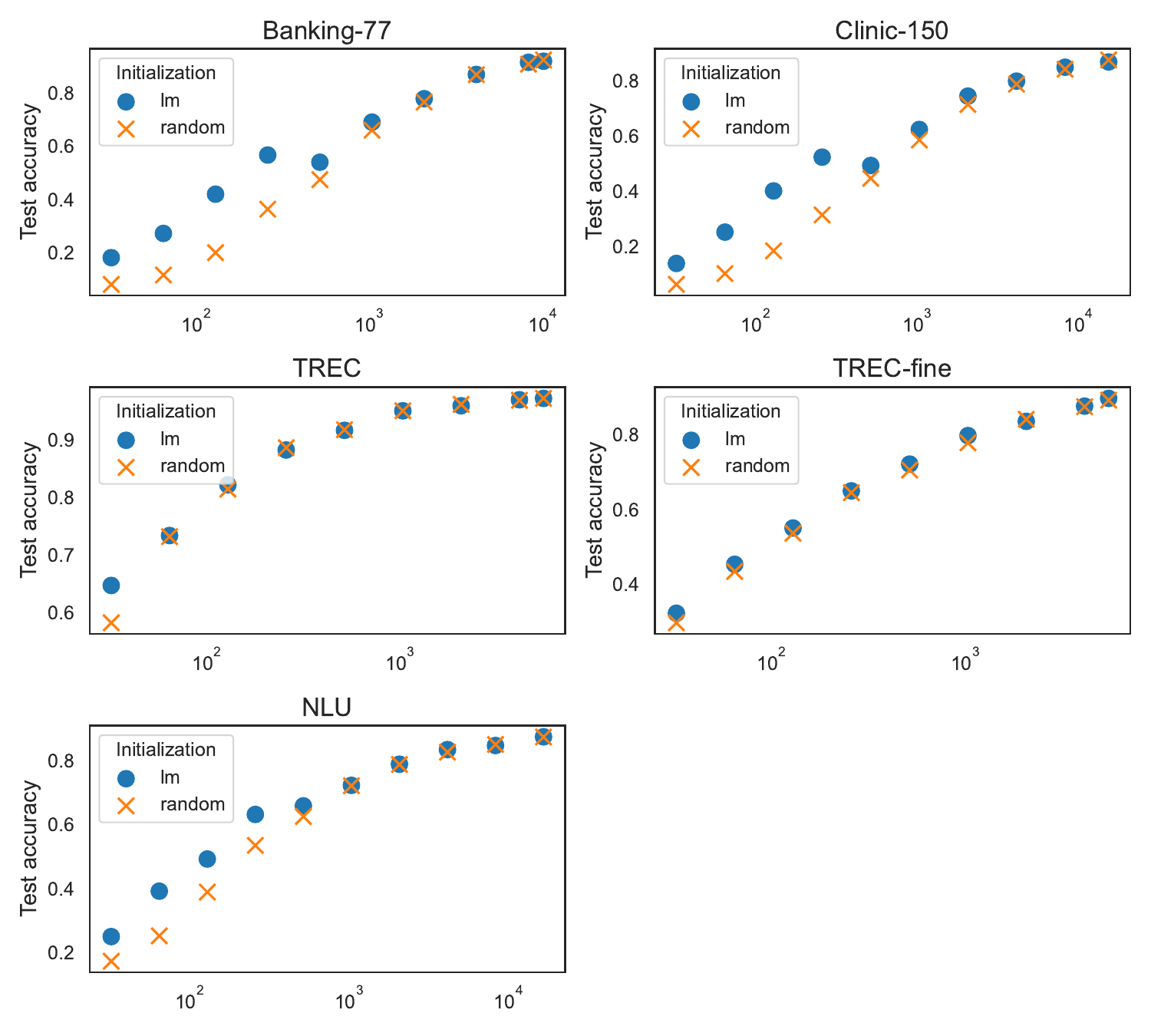}
\caption{Comparing initialization methods of the classification head when finetuning a PEFT llama-2-7b model. Averaged (best) test accuracy over 5 random seeds. Initialization with \emph{lm} subsamples the pretrained language-modeling head at the first token of the target label, while random samples random weights.}
\label{fig:lm-init}
\end{figure}

\paragraph{Initialization of the classification head} While in our default setting we initialize the classification head from the pretrained LM head, subsampled at the representation of the first token in each label, we investigate the efficacy of this approach by contrasting with a randomly initialized classification head. \Cref{fig:lm-init} shows that while in the few-shot regime, this approach has significant advantage, as the training set grows in size the difference shrinks to become negligible. In no case was random initialization better than this approach.

\begin{figure}
  \centering
  \includegraphics[width=\linewidth]{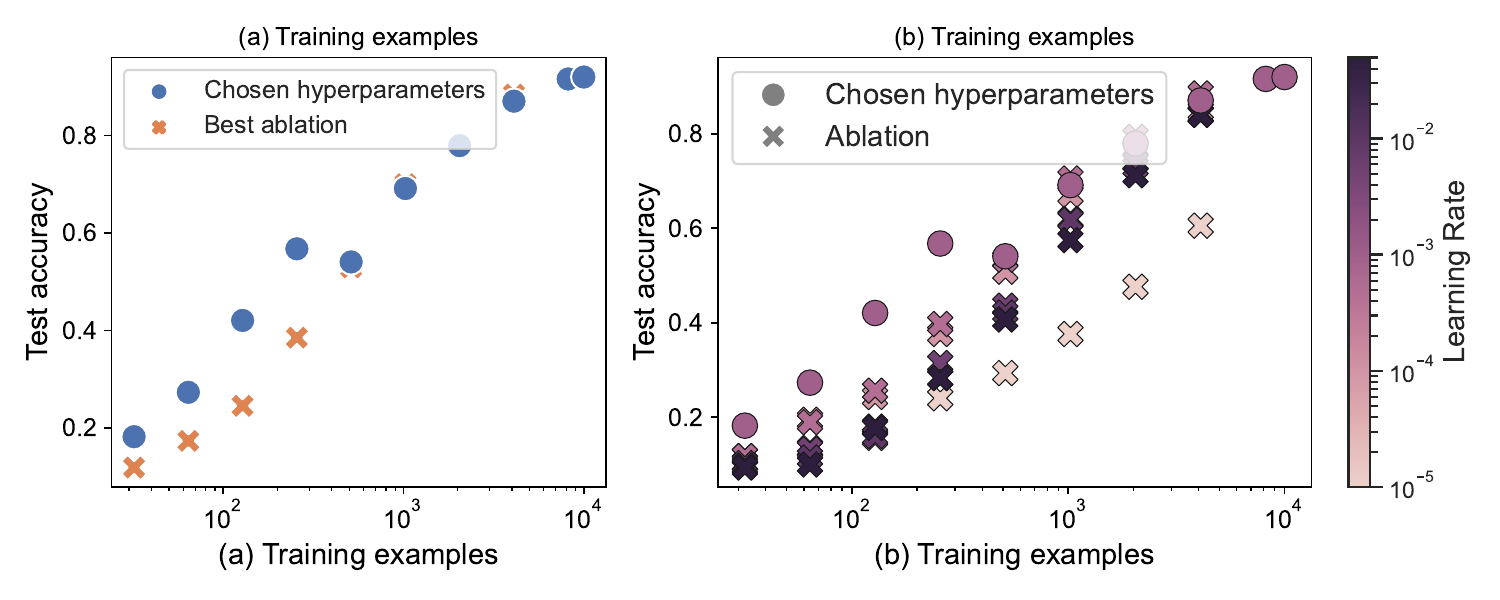}
\caption{Comparing hyperparameters when finetuning a PEFT llama-2-7b model on \banking. Averaged (best) test accuracy over 3 random seeds. (a) Comparing our fixed LoRA configurations to the best alternative configuration (at each scale) we tried. (b) Comparing different learning rates.}
\label{fig:hp-tuning}
\end{figure}

\paragraph{Hyperparameter tuning} To remain comparable in terms of compute efficient finetuning, we did not perform extensive hyper-parameter tuning per task, and instead experimented with a good global setting on a single dataset (\banking). Specifically, we experimented with different learning rates, different LoRA ranks ($r$) and $\alpha$ \citep{hu2021lora} and also tried applying RSlora \citep{Kalajdzievski2023ARS} which sets the scaling factor to $\frac{\alpha}{\sqrt{r}}$ as some evidence suggest it can outperform the original method. \Cref{fig:hp-tuning} summarizes the results, depicting average test accuracy against training examples with different settings.

Ultimately, we found that using HuggingFace's \citep{wolf2020transformers} default parameters of $r=8$, $\alpha=32$, LoRA dropout of $0.1$ and a learning rate of $1e-3$ to work best. In all cases, we used batch sizes of 32 and weight decay of $0.01$.

It is possible that methods specialized for finetuning in small-data regimes, such as T-few \cite{liu2022fewshot}, might close the gap between ICL and PEFT in the small-data regimes. 
We did not consider T-few in our analysis because of its additional pretraining stage, which imposes substantial additional cost, and because T-few was developed with a focus on encoder-decoder models and we consider only decoder-only models in our setting.

\newpage
\section{Block attention patterns}
\label{appendix:block-attn-ablations}

We aim to construct a modified attention pattern where demonstrations can only attend to a limited block of local demonstrations.

 The most naive strategy is to only allow attention within a local block (Figure~\ref{fig:naive-block}). Other variants relax the mask to allow attention to the first block, which acts as an attention sink \citep{xiao2024efficient} (Figure~\ref{fig:sink-block}), or allowing attention back to the immediately prior local block(s) (as seen as part of \citet{acharya2024starattentionefficientllm})(Figure~\ref{fig:local-block}).
\begin{figure}[h]

  \centering

\begin{subfigure}{0.2\linewidth}
  \centering
        \begin{tikzpicture}[scale=0.22]    
        \def\N{15}  % Number of tokens

        % Causal attention pattern
        \foreach \i in {1,...,\N} {
            \foreach \j in {1,...,\i} {
                \fill[Cyan!30] (\j, -\i) rectangle ++(1,-1);
            }
        }
        
        % Full attention for last token
        \foreach \j in {1,...,\N} {
            \fill[Cyan!30] (\j, -\N) rectangle ++(1,-1);
        }
        
        % Dark blue diagonal
        \foreach \i in {1,...,\N} {
            \fill[NavyBlue!60] (\i, -\i) rectangle ++(1,-1);
        }

        % Draw the grid
        \foreach \i in {1,...,\N} {
            \foreach \j in {1,...,\N} {
                \draw[gray!70] (\i, -\j) rectangle ++(1,-1);
            }
        }
                
            \end{tikzpicture}

    \caption{Causal attention}
    \label{fig:causal-attn}
    \end{subfigure}
     \hspace{0.03\linewidth}
\begin{subfigure}{0.2\linewidth}
  \centering
        \begin{tikzpicture}[scale=0.22]    
        \def\N{15}  % Number of tokens
        \def\B{2}   % Block size

        % Full attention for last block
        \foreach \j in {1,...,\N} {
            \fill[Cyan!30] (\j, -\N) rectangle ++(1,-1);
        }
    
    % Dark blue diagonal
        \foreach \i in {1,...,\N} {
            \fill[NavyBlue!60] (\i, -\i) rectangle ++(1,-1);
        }

        % Draw the grid
        \foreach \i in {1,...,\N} {
            \foreach \j in {1,...,\N} {
                \draw[gray!70] (\i, -\j) rectangle ++(1,-1);
            }
        }
        \end{tikzpicture}

    \caption{Only the current block}
    \label{fig:naive-block}
    \end{subfigure} 
            \hspace{0.03\linewidth}
        \begin{subfigure}{0.2\linewidth}
  \centering
        \begin{tikzpicture}[scale=0.22]    
        \def\N{15}  % Number of tokens
        \def\B{2}   % Block size

        % Full attention for last block
        \foreach \j in {1,...,\N} {
            \fill[Cyan!30] (\j, -\N) rectangle ++(1,-1);
        }

        % local blocks
        \foreach \i in {2,...,\N} {
            \fill[Cyan!30] (\i, -\i) rectangle ++(-1,-1);
        }

    % Dark blue diagonal
        \foreach \i in {1,...,\N} {
            \fill[NavyBlue!60] (\i, -\i) rectangle ++(1,-1);
        }

        % Draw the grid
        \foreach \i in {1,...,\N} {
            \foreach \j in {1,...,\N} {
                \draw[gray!70] (\i, -\j) rectangle ++(1,-1);
            }
        }
        \end{tikzpicture}

    \caption{One local block}
    \label{fig:local-block}
    \end{subfigure} \\
    \begin{subfigure}{0.20\linewidth}
  \centering
        \begin{tikzpicture}[scale=0.22]    
        \def\N{15}  % Number of tokens
        \def\B{2}   % Block size

          % Sink attention pattern
        \foreach \i in {1,...,\N} {
            \fill[Cyan!30] (1, -\i) rectangle ++(1,-1);
        }

        % Full attention for last block
        \foreach \j in {1,...,\N} {
            \fill[Cyan!30] (\j, -\N) rectangle ++(1,-1);
        }
    
    % Dark blue diagonal
        \foreach \i in {1,...,\N} {
            \fill[NavyBlue!60] (\i, -\i) rectangle ++(1,-1);
        }

        % Draw the grid
        \foreach \i in {1,...,\N} {
            \foreach \j in {1,...,\N} {
                \draw[gray!70] (\i, -\j) rectangle ++(1,-1);
            }
        }
        \end{tikzpicture}
    \caption{Attention sink}
    \label{fig:sink-block}
    \end{subfigure} 
            \hspace{0.03\linewidth}
        \begin{subfigure}{0.2\linewidth}
  \centering
        \begin{tikzpicture}[scale=0.22]    
        \def\N{15}  % Number of tokens
        \def\B{2}   % Block size

        % Full attention for last block
        \foreach \j in {1,...,\N} {
            \fill[Cyan!30] (\j, -\N) rectangle ++(1,-1);
        }

         % Sink attention pattern
        \foreach \i in {1,...,\N} {
            \fill[Cyan!30] (1, -\i) rectangle ++(1,-1);
        }

        % local blocks
        \foreach \i in {2,...,\N} {
            \fill[Cyan!30] (\i, -\i) rectangle ++(-1,-1);
        }

    % Dark blue diagonal
        \foreach \i in {1,...,\N} {
            \fill[NavyBlue!60] (\i, -\i) rectangle ++(1,-1);
        }

        % Draw the grid
        \foreach \i in {1,...,\N} {
            \foreach \j in {1,...,\N} {
                \draw[gray!70] (\i, -\j) rectangle ++(1,-1);
            }
        }
        \end{tikzpicture}

    \caption{StarAttention}
    \label{fig:staratnn-block}
    \end{subfigure}
            \hspace{0.03\linewidth}
    \begin{subfigure}{0.2\linewidth}
  \centering
        \begin{tikzpicture}[scale=0.22]    
        \def\N{15}  % Number of tokens
        \def\B{2}   % Block size

          % Sink attention pattern
        \foreach \i in {1,...,\N} {
            \fill[Cyan!30] (1, -\i) rectangle ++(1,-1);
        }

        % local blocks
        \foreach \i in {3,...,\N} {
            \fill[Cyan!30] (\i, -\i) rectangle ++(-1,-1);
        }

        \foreach \i in {4,...,\N} {
            \fill[Cyan!30] (\i, -\i) rectangle ++(-2,-1);
        }

        % Full attention for last block
        \foreach \j in {1,...,\N} {
            \fill[Cyan!30] (\j, -\N) rectangle ++(1,-1);
        }
    
    % Dark blue diagonal
        \foreach \i in {1,...,\N} {
            \fill[NavyBlue!60] (\i, -\i) rectangle ++(1,-1);
        }

        % Draw the grid
        \foreach \i in {1,...,\N} {
            \foreach \j in {1,...,\N} {
                \draw[gray!70] (\i, -\j) rectangle ++(1,-1);
            }
        }
        \end{tikzpicture}

    \caption{Our chosen attention}
    \label{fig:chosen-block}
    \end{subfigure}
    \\
    \caption{Representative examples of attention masks we considered. In each diagram, the squares represent a block of $b$ examples each, \textit{not} individual tokens.}
\end{figure}

As a test of these methods, we perform block attention with each attention mask on \banking with \longllama, with the block size $b=50$ and 500 demonstrations in-context.

\begin{table*}[h]
\begin{center}
\begin{tabular}{lllll}
\toprule
\multicolumn{1}{c}{\bf Diagram}  & \multicolumn{1}{c}{\bf Block size $b$}  &\multicolumn{1}{c}{\bf Sink blocks}  &\multicolumn{1}{c}{\bf Local blocks}  &\multicolumn{1}{c}{\bf Accuracy}  
\\ 
\midrule 
Figure~\ref{fig:causal-attn} & 500 (full causal) & - & - & 80.04 \\
\midrule
Figure~\ref{fig:naive-block} & 50 & 0 & 0 & 18.72 \\
Figure~\ref{fig:local-block} & 50 & 0 & 1 & 51.16 \\
Figure~\ref{fig:sink-block} & 50 & 1 & 0 & 62.15 \\
Figure~\ref{fig:staratnn-block} & 50 & 1 & 1 & 75.12 \\
Figure~\ref{fig:chosen-block} & 50 & 1 & 2 & \textbf{77.51} \\
\bottomrule
\end{tabular}
\end{center}
\caption{Comparison of attention strategies and performance.}
\label{tab:results-main}

\end{table*}
Neither a sink block nor local attention alone recovers close to full attention performance, although both are substantial improvements over the naive strategy. The best performance is achieved with both sink and local blocks; although one local block is sufficient to recover 94\% of the performance of the full attention method, we choose two local blocks to minimize performance degradation. 
As a result, we use the attention pattern in Figure~\ref{fig:chosen-block} for the experiments in the paper. 
\newpage
\section{Prompt formatting and examples from datasets}
\label{appendix:prompts}
As a demonstration of the datasets, we provide an example of 3-shot prompting for each dataset with the prompt formatting we used (and with examples drawn from the training set of each dataset).

Prompt formatting and instruction phrasing can have significant impact on performance \citep{sclar2023quantifying}; we keep the formatting consistent with prior work \citep{Ratner2022ParallelCW}, with prefixes for the input and output for each exemplar. Because we use predominately non-instruction-tuned models, we do not add an additional instruction or system prompt.

\subsection{\trec}

\textbf{TREC} \citep{hovy-etal-2001-toward, li-roth-2002-learning} is a question classification dataset with two granularities of labels. We refer to the 6-label coarse classification as \trec. \\

\begin{figure}[h!]
\centering
 \includegraphics[height=1.5cm]{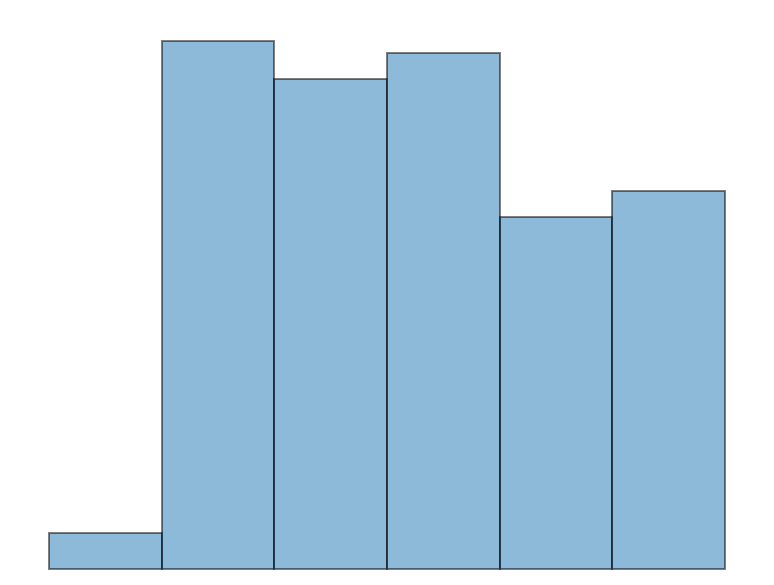} 
\caption{The label distribution of \trec. One label is much less frequent than the rest.} 
\end{figure} 

\begin{tcolorbox}[colback=gray!10!white, colframe=gray!50!black, boxrule=0.5pt, sharp corners]
    \textit{Question: How does light travel through the void of space if there is no medium for it to ` wave ' or ` pulse ' . \\ Type: description \\ == \\ Question: What cathedral was Thomas Becket murdered in ? \\ Type: location \\ == \\ Question: The lawyer who represented Randy Craft , what was his name ? \\ Type: human \\ == \\ Question: What are the rites accompanying the circumcision of a newly born-child in Judaism called ? \\ Type:}
\captionof{figure}{Example 3-shot prompt for \trec.}

\end{tcolorbox}

\subsection{\trecfine}
We refer to \trec's 50-label finegrained classification as \trecfine. \\

\begin{figure}[h!]
\centering
 \includegraphics[width=5cm]{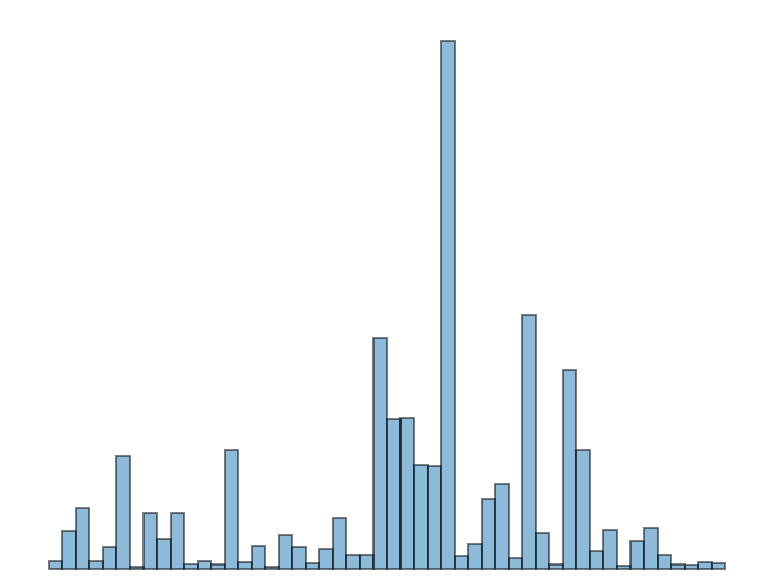} 
\caption{The label distribution of \trecfine.} 
\end{figure} 

\begin{tcolorbox}[colback=gray!10!white, colframe=gray!50!black, boxrule=0.5pt, sharp corners]
    \textit{Question: How does light travel through the void of space if there is no medium for it to ` wave ' or ` pulse ' . \\ Type: description manner \\ == \\ Question: What cathedral was Thomas Becket murdered in ? \\ Type: location other \\ == \\ Question: The lawyer who represented Randy Craft , what was his name ? \\ Type: human individual \\ == \\ Question: What are the rites accompanying the circumcision of a newly born-child in Judaism called ? \\ Type: }
\captionof{figure}{Example 3-shot prompt for \trecfine.}
\end{tcolorbox}

\subsection{ \nlu}

\begin{figure}[h!]
\centering
 \includegraphics[height=5cm]{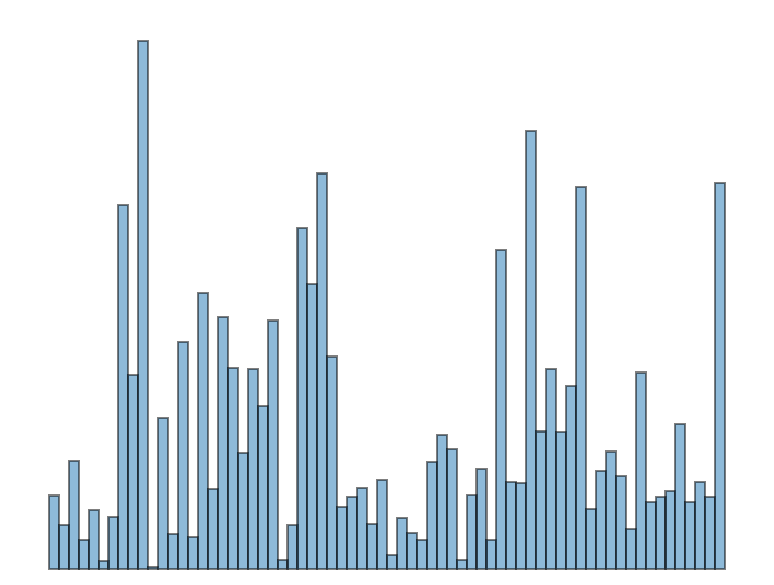} 
\caption{The label distribution of \nlu.} 
\end{figure} 

\textbf{NLU} \citep{nlu} is a 68-way intent classification dataset in the conversational domain. The original paper evaluates on 64 of the intents; we use all 68. The data is licensed under CC BY 4.0.
\begin{tcolorbox}[colback=gray!10!white, colframe=gray!50!black, boxrule=0.5pt, sharp corners]
    \textit{utterance: oh it is nice one, olly. \\ intent: general praise \\ == \\ utterance: nope wrong. \\ intent: general negate \\ == \\ utterance: what events near hear are happening this week \\ intent: recommendation events \\ == \\ utterance: play fishing podcasts that are favorited \\ intent: }
\captionof{figure}{Example 3-shot prompt for \nlu.}
\end{tcolorbox}

\subsection{\banking}

\begin{figure}[h!]
\centering
 \includegraphics[height=5cm]{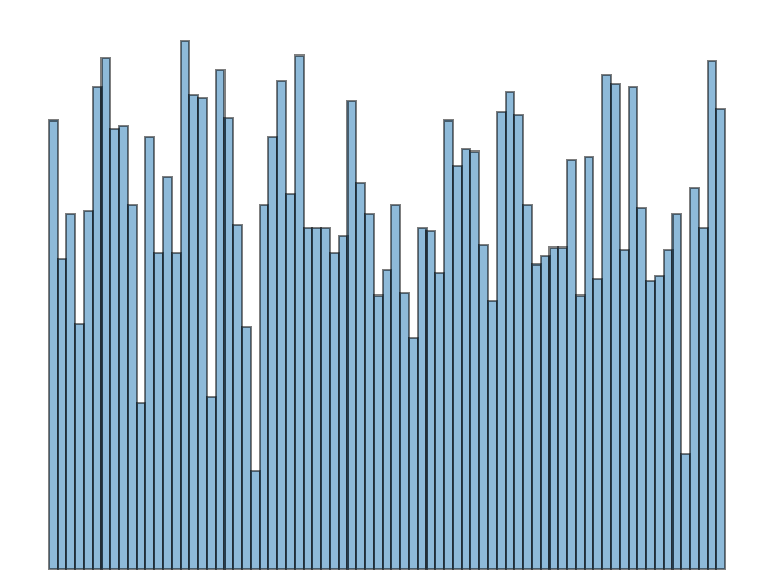} 
\caption{The label distribution of \banking.} 
\end{figure} 

\textbf{\banking} \citep{casanueva-etal-2020-efficient} is a 77-way intent classification task in the financial domain. Although the accuracy of some labels in BANKING77 has been criticized \citep{ying-thomas-2022-label}, we report results here on the original dataset for consistency with prior work. The data is licensed under CC BY 4.0.

\begin{tcolorbox}[colback=gray!10!white, colframe=gray!50!black, boxrule=0.5pt, sharp corners]
    \textit{query: How long will my payment be pending? \\ intent: pending card payment \\ == \\ query: My physical card is not working \\ intent: card not working \\ == \\ query: i cant seem to activate card \\ intent: activate my card \\ == \\ query: I didn't set up a direct debit payment on my account. \\ intent: }
\captionof{figure}{Example 3-shot prompt for \banking.}
\end{tcolorbox}

\subsection{\clinic}

\begin{figure}[h!]
\centering
 \includegraphics[height=5cm]{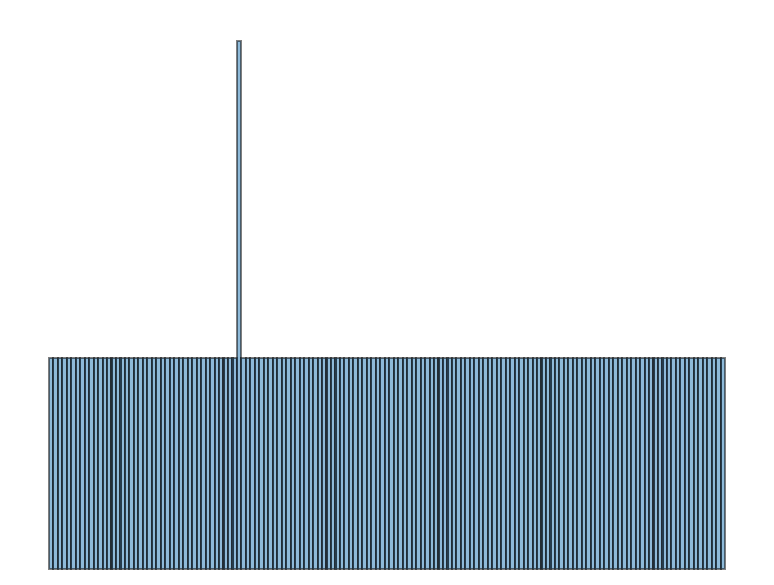} 
\caption{The label distribution of \clinic. It is balanced except for the ``out of scope'' label, which has additional data points in the split we use.} 
\end{figure}

\textbf{\clinic} \citep{clinic150} is a 151-way, multi-domain intent classification task; examples are either labeled with an intent from one of 10 domains or with the catch-all ``out-of-scope`` label. We use the ``plus`` train split from the original paper, which adds additional ``out-of-scope'' examples to the dataset. The data is licensed under CC BY 3.0.

\begin{tcolorbox}[colback=gray!10!white, colframe=gray!50!black, boxrule=0.5pt, sharp corners]
    \textit{utterance: how much is my comcast bill \\ intent: bill balance \\ == \\ utterance: tell me about yourself \\ intent: what is your name \\ == \\ utterance: how to build up my credit score \\ intent: improve credit score \\ == \\ utterance: are you employed by me \\ intent:}
\captionof{figure}{Example 3-shot prompt for \clinic.}
\end{tcolorbox}

\subsection{\samsum}
\textbf{\samsum} \citep{Gliwa_2019} is a text message summarization corpus; both the text conversations and summaries were written by linguists. The conversations can be two or more participants, and may contain emojis, emoticons, or tags that indicate the use of an image or GIF; the summaries are generally 1-3 sentences long. The data is licensed under CC BY-NC-ND 4.0.
\newpage 

\begin{tcolorbox}[colback=gray!10!white, colframe=gray!50!black, boxrule=0.5pt, sharp corners]
    \textit{Conversation: Zara: <file\_gif> \\
Zara: something terrible happened :( \\
Stanley: what? You OK? \\
Zara: yes, I'm fine! I went to the swimming pool \\
Zara: and lost the earring you gave me for birthday :( \\
Stanley: oh Jesus, I thought something bad happened to you! \\
Summary: Zara has lost the earring, which Stanley gave her for birthday, when she was at the swimming pool.
 \\ == \\
 Conversation: Rose: hey congratulations for the baby boy! i wish him all the health and happiness in life. \\
Ela: thank you so much for the wishes. \\
Rose: your welcome.. so hows he and you? all ok \\
Ela: yes everything is great thanks \\
Rose: will come to see you and the baby boy \\
Ela: Sure will wait to see you.. \\
Summary: Ela just gave birth to a boy. Rose will visit them shortly. \\ == \\
Conversation: Kim: Are you going to the conference in SF? \\
Jenny: I should, I know, it would be good for my career \\
Jeff: no, not so much, I think it's bullshit that it's so important \\
Simon: is it? \\
Jeff: sure, the whole net-working thing doesn't really matter, I think \\
Jeff: nobody offers you a job at a conference\\
Jeff: and it costs so much to fly to SF\\
Kim: I would like to go also to see what's going on in the field
Kim: to meet people, see new trends, ideas \\
Kim: I think it's important for an academic \\
Jeff: this may be true, if you can afford \\
Kim: the flight is about €500, right? \\
Simon: true \\
Jeff: and then more money for accommodation \\
Jeff: it can easily pile up to €2000 \\
Kim: you're quite right, unfortunately \\
Jeff: because it also doesn't make sense to fly to California for 3 days \\
Jeff: it would be also extremely disturbing, with the jet lag etc. \\
Kim: you're so right :( \\
Jeff: so think about it first \\
Summary: Jeff is not going to the conference in SF. The flight is expensive. \\ == \\
Conversation: Amanda: I baked  cookies. Do you want some? \\
Jerry: Sure! \\
Amanda: I'll bring you tomorrow :-) \\
Summary: 
 }
\captionof{figure}{Example 3-shot prompt for \samsum.}
\end{tcolorbox}

\newpage
\section{Additional details}
\label{appendix:extras}
\subsection{Models selected}
The majority of the analysis in the paper concerns these five models: 
\begin{enumerate}[nosep]
    \item \textbf{Llama2} \citep{touvron2023llama} is a decoder-only model trained with a 4096 context length. We use the non-instruct (non-chat) variant because it is more commonly used as a base model for long-context finetuning and because we observed very similar performance between the chat and non-chat variants in our initial experiments. 
    \item \textbf{Llama2-32k} \citep{TogetherAI_2023} is a version of Llama-2-7b finetuned by TogetherAI for a 32k context window. We use the non-instruct version. 
    \item \textbf{Llama2-80k} \citep{fu2024data} is a version of Llama-2-7b finetuned with 80k context and a carefully designed long-document data mixture.
    \item \textbf{Mistral-7b-v0.2} \citep{jiang2023mistral}. is the instruct version of Mistral-v0.2 (the non-instruct model is not publicly available). The trained context length of Mistral-7B-Instruct-v0.2 is 32k tokens. 
    \item \textbf{Qwen2.5-7B} \citep{qwen2.5} is a decoder-only model with multilingual support. The trained context length of Qwen2.5-7B is 128k tokens. 
\end{enumerate}

While all of these models can extrapolate to inputs longer than their trained context length, we restrict the lengths of inputs to fit within the trained context length; this represents the best case performance without the additional confound of the extrapolation strategy. 

\subsection{Computational cost}
For our finetuning experiments, we used approximately 50 GPU-days of compute on 80GB A100 GPUs. The computational cost of the in-context learning experiments was approximately 75 GPU-days, primarily using 48GB A6000 GPUs. All experiments were run on local clusters (i.e. not using cloud providers), with the exception of the frontier models experiments, which used Together AI's API (for Llama 3.1-405B) and the Anthropic API (for Claude). 

\end{document}